\documentclass[lettersize,journal]{IEEEtran}
\usepackage{amsmath,amsfonts}
\usepackage{enumitem}
\usepackage{algorithm}
\usepackage{algpseudocode} 
\usepackage{array}
\usepackage[caption=false,font=normalsize,labelfont=sf,textfont=sf]{subfig}
\usepackage{textcomp}
\usepackage{stfloats}
\usepackage{url}
\usepackage{verbatim}
\usepackage{graphicx}
\usepackage{cite}
\usepackage{amsmath}
\usepackage[colorlinks=true, 
            linkcolor=blue, 
            citecolor=blue, 
            urlcolor=blue]{hyperref}

\usepackage{booktabs}
\usepackage{multirow}
\usepackage{siunitx}
\usepackage[table]{xcolor}   

\begin{document}

\title{PrefO2O: Zero-Shot Preference-Based Reinforcement Learning \\ via Offline-to-Online Latent Preference Transfer}

\author{
Ruiqi Wang$^{1\dagger}$,
Dezhong Zhao$^{2\dagger}$,
Xuesu Xiao$^{3}$,
Erdem B{\i}y{\i}k$^{1}$,
Byung-Cheol Min$^{4}$,
and Guohua Chen$^{2}$
\thanks{$\dagger$ Equal Contribution}
\thanks{$^{1}$ University of Southern California, Los Angeles, CA, USA.}
\thanks{$^{2}$ Beijing University of Chemical Technology, Beijing, China.}
\thanks{$^{3}$ George Mason University, Fairfax, VA, USA.}
\thanks{$^{4}$ Indiana University Bloomington, Bloomington, IN, USA.}
}

\markboth{Journal of \LaTeX\ Class Files,~Vol.~14, No.~8, August~2021}%
{Shell \MakeLowercase{\textit{et al.}}: A Sample Article Using IEEEtran.cls for IEEE Journals}


\maketitle

\begin{abstract}
Preference-based reinforcement learning (PbRL) has emerged as a promising paradigm for learning robot behaviors aligned with human intent. However, its scalability is fundamentally limited by the need to collect expensive human preference labels for every new task. In this work, we present PrefO2O, an offline-to-online latent preference transfer framework that enables zero-shot PbRL on unseen target tasks without requiring any human feedback. The core idea is to learn a task-invariant latent preference space from a small amount of offline preference data on source tasks, where transferable preference structures are disentangled from task-specific variations. This learned manifold allows PrefO2O to transfer preference structures and reward priors to new tasks under distribution shift, while confidence-weighted optimal transport alignment generates reliable pseudo-preference supervision for online reward adaptation. Extensive experiments on robotic manipulation and locomotion benchmarks demonstrate that PrefO2O substantially outperforms existing preference-transfer baselines and matches or exceeds PbRL methods trained with real human feedback, while generalizing across multi-to-one and one-to-one transfer settings as well as across different robot embodiments. More information is available at our  \href{https://zdz0086.github.io/PrefO2O_IROS26/}{\text{project website}}.
\end{abstract}


\section{Introduction}

\IEEEPARstart{R}{einforcement} learning (RL) has achieved remarkable success in robotics~\cite{kober2013reinforcement}, yet its dependence on carefully engineered reward functions requires substantial domain expertise and remains vulnerable to reward hacking~\cite{gupta2022unpacking}. Preference-based RL (PbRL) offers a compelling alternative: instead of specifying rewards explicitly, it learns reward models from human comparative feedback, thereby aligning robot behaviors with human intent~\cite{christiano2017deep,lee2021b,baraka2026human,wang2024personalization}.

However, PbRL introduces a fundamental scalability bottleneck. For each new task, high-quality human preference labels must typically be collected from scratch to train an effective reward model~\cite{christiano2017deep,lee2021pebble,biyik2024active}. This per-task annotation overhead limits the practical deployment of PbRL in real-world robotic systems, where robots are expected to generalize rapidly across diverse tasks, embodiments, and operating conditions.

Several directions have been explored to reduce this scalability bottleneck. One line of work leverages foundation models to generate synthetic preference labels, enabling scalable feedback without direct human annotation~\cite{wang2025prefclm,wang2024rl,liu2025vlp,wang2025primt}. However, synthetic feedback might be inconsistent, diverge from genuine human intent, and often incurs substantial computational cost~\cite{huang2025survey}. Another line of work applies meta-learning to PbRL, enabling faster adaptation to new tasks by meta-training reward models across task distributions~\cite{hejna2023few,finn2017model}. Yet these methods require extensive preference data during meta-training and still demand target-task feedback, which remains impractical in many real-world robotic settings.

More recent work has investigated zero-shot preference transfer by propagating pseudo-preference labels from source tasks to target tasks through trajectory alignment, for example using optimal transport~\cite{liu2024pearl}. While promising, existing methods typically operate directly on shallow trajectory-level similarity metrics. Such alignment can be brittle under distribution shifts, especially when tasks differ in dynamics, objectives, or state-action dimensionality. Moreover, treating preference transfer primarily as offline label propagation provides limited guidance for reward adaptation and policy improvement on the target task.

\begin{figure*}[t]
    \centering
    \includegraphics[width=0.98\linewidth]{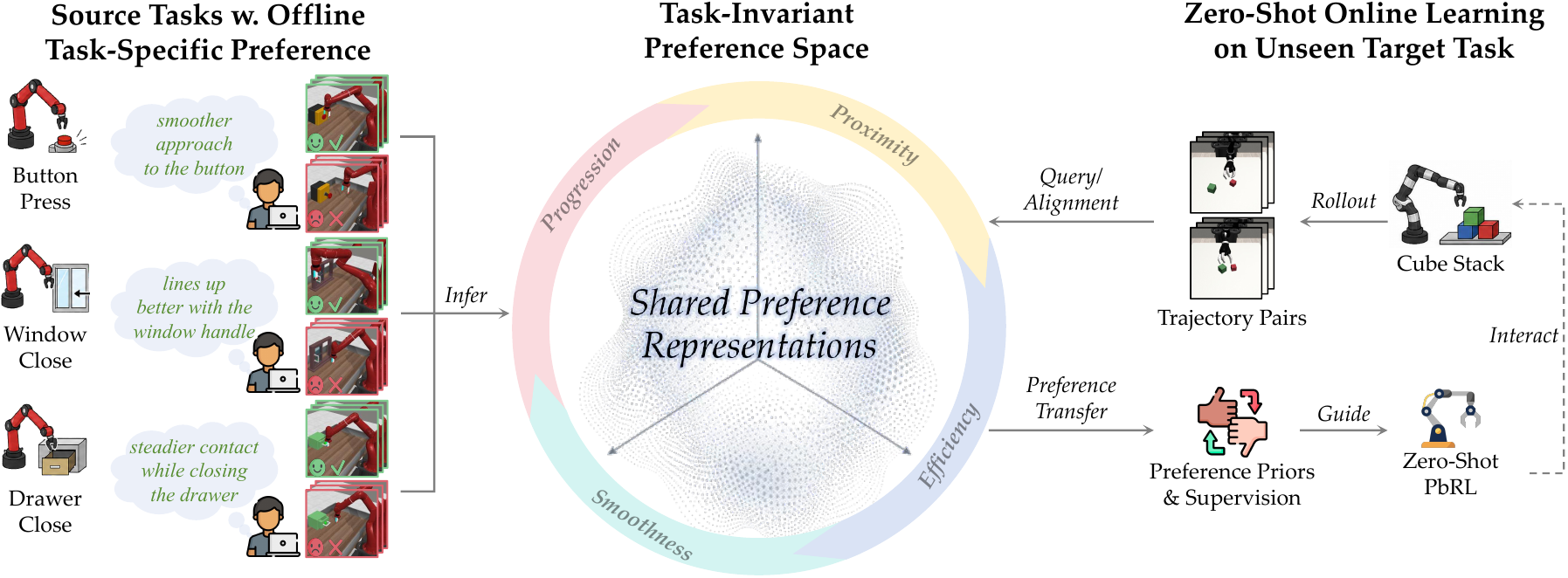}
    \caption{Conceptual illustration of PrefO2O. Offline preference data from multiple source tasks are used to learn a task-invariant latent preference space that captures shared preference structures while absorbing task-specific variations. When deployed on an unseen target task, this learned space enables both transferable reward priors and OOD-aware pseudo-preference supervision for zero-shot online PbRL without target-task human feedback.}
    \label{fig:concept}
    \vspace{-15pt}
\end{figure*}

In this work, we explore how to enable PbRL to transfer across tasks without collecting human feedback on each new target task. Our key insight is that, although task objectives and dynamics may vary, human preferences often contain shared structures that are transferable across related tasks. For example, across manipulation tasks with distinct goals, humans may consistently favor efficient approaches, penalize unnecessary motion, and prefer stable object interactions. These recurring preference patterns suggest that reward learning need not start from scratch for every new task.

Motivated by this insight, we propose PrefO2O, an offline-to-online latent preference transfer framework for zero-shot PbRL across tasks. As illustrated in Fig.~\ref{fig:concept}, PrefO2O learns a task-invariant latent preference space from offline preference data collected across multiple source tasks, abstracting away task-specific variations to expose shared, transferable preference structures. Upon deployment to an unseen target task, PrefO2O operates in this learned latent space to transfer preference priors and generate reliable pseudo-preference supervision, enabling online PbRL under distribution shift and heterogeneous state-action spaces. As overviewed in Fig.~\ref{fig:framework}, PrefO2O consists of two stages:

\textit{Offline Phase.} We jointly train a trajectory encoder, a preference encoder, and a meta reward model on source-task preference data. The trajectory encoder maps raw, potentially heterogeneous trajectories into compact behavioral representations. The preference encoder learns a structured latent preference manifold over pairwise trajectory differences, capturing shared preference structures across tasks. Concurrently, the meta reward model learns to map trajectory representations to preference-aligned scalar rewards, providing a transferable reward prior for target-task learning.

\textit{Online Phase.} The pretrained trajectory encoder and reward model serve as a warm start for the target task. The preference encoder remains frozen to preserve the learned manifold, which serves as a stable reference for OOD-aware preference alignment via optimal transport. This alignment generates confidence-weighted pseudo-preference labels that guide continued reward adaptation and policy optimization on the target task.

Our key contributions are as follows:
\begin{itemize}[leftmargin=*]
\item We propose PrefO2O, an offline-to-online latent preference transfer framework that enables zero-shot PbRL across tasks, eliminating the need for human preference feedback on new tasks.
\item We introduce a task-invariant latent preference space that abstracts away task-specific variation to capture shared, transferable preference structures. PrefO2O leverages this learned space to transfer preference priors and generate reliable pseudo-preference supervision for unseen target tasks via OOD-aware optimal transport alignment.
\item We validate PrefO2O on two manipulation and one locomotion benchmarks, demonstrating superior performance over existing preference transfer baselines and PbRL methods trained with real human labels or synthetic feedback. Comprehensive ablation studies confirm the contribution of each component, and we show that PrefO2O generalizes across multi-to-one and one-to-one transfer settings, as well as both intra- and cross-embodiment scenarios.
\end{itemize}

\section{Background and Preliminaries}

\subsection{Preference-Based Reinforcement Learning}

PbRL aims to learn a reward model from human comparative feedback over trajectories~\cite{christiano2017deep,hwang2026causally,wang2026preference,peng2024pragmatic,torne2026freeform,brown2019extrapolating,wilde2020active}. Compared with standard RL, which often requires carefully engineered reward functions, PbRL alleviates the burden of reward specification and improves behavioral alignment with human intent. Preference feedback is also often easier to provide than demonstrations, since it requires only relative judgments between trajectory segments rather than precise action-level guidance~\cite{yuan2026prefmoe,myers2023active}.

Formally, a trajectory segment is a finite-horizon sequence of state-action pairs $\sigma = \{(s_t, a_t)\}_{t=1}^{H}$. Given a pair $(\sigma_A, \sigma_B)$, a human annotator provides a preference label $\Upsilon \in \{0, 1, -1\}$, where $\Upsilon{=}1$ indicates $\sigma_A \succ \sigma_B$, $\Upsilon{=}0$ indicates $\sigma_B \succ \sigma_A$, and $\Upsilon{=}{-1}$ denotes indifference $(\sigma_A \sim \sigma_B)$.

A parametric reward model $r_\psi(s,a)$ assigns a scalar reward to each state-action pair. The cumulative reward over a trajectory segment is $R_\psi(\sigma) = \sum_{t=1}^{H} r_\psi(s_t, a_t)$. Under the Bradley--Terry model~\cite{bradley1952rank}, the probability that $\sigma_A$ is preferred over $\sigma_B$ is
\begin{equation}
  P_\psi[\sigma_A \succ \sigma_B]
    = \frac{\exp R_\psi(\sigma_A)}
           {\exp R_\psi(\sigma_A)
            + \exp R_\psi(\sigma_B)}
  \label{eq:bt_model}
\end{equation}

The reward model parameters $\psi$ are optimized by minimizing the cross-entropy loss over a preference dataset $\mathcal{D}_{\text{pref}} = \{(\sigma_{A,i},\,\sigma_{B,i},\,\Upsilon_i)\}_{i=1}^{N}$:
\begin{equation}
\begin{split}
  \mathcal{L}_{\text{pref}}(\psi)
    = -\mathbb{E}_{(\sigma_A,\sigma_B,\Upsilon)\sim\mathcal{D}_{\text{pref}}}
      \Bigl[
        &\mathbb{I}(\Upsilon{=}1)\log P_\psi[\sigma_A \succ \sigma_B]\\
        +\;&\mathbb{I}(\Upsilon{=}0)\log P_\psi[\sigma_B \succ \sigma_A]
      \Bigr]
\end{split}
  \label{eq:pref_loss}
\end{equation}
where $\mathbb{I}(\cdot)$ is the indicator function.

\subsection{Preference Transfer in PbRL}

Standard PbRL methods typically learn reward models from scratch for each new task, incurring substantial annotation costs that limit scalability in real-world settings. Preference transfer seeks to mitigate this bottleneck by reusing preference knowledge from previously seen source tasks to reduce, or ideally eliminate, feedback collection on new target tasks.

Hejna et al.~\cite{hejna2023few} take an initial step in this direction by integrating PbRL with MAML~\cite{finn2017model}, meta-training a reward model across multiple tasks for few-shot adaptation. While effective, this approach remains data-intensive at both stages: meta-training requires thousands of labeled queries per source task across about $10$ tasks, and adaptation still demands hundreds to thousands of additional queries on the target task. In contrast, our approach operates with modest preference budgets on source tasks and requires zero human feedback on the target task, while generalizing across both many-to-one and one-to-one transfer settings.

The work most closely related to ours is PEARL~\cite{liu2024pearl}, which enables zero-shot preference transfer by aligning source and target trajectory distributions via optimal transport and propagating pseudo-preference labels through the resulting transport matrix. However, operating directly in the raw state-action space can make the alignment sensitive to distribution shifts across tasks, where trajectory-level similarity metrics may become unreliable. In addition, PEARL assumes a shared state-action dimensionality across tasks, which limits transfer across different embodiments or sensing configurations. Moreover, it treats preference transfer primarily as label propagation, rather than learning a transferable preference representation that can support online reward and policy adaptation.

In contrast, PrefO2O performs offline-to-online preference transfer through a task-invariant latent preference space that absorbs task-specific variation. This learned space enables OOD-aware optimal transport for pseudo-preference generation under distribution shift, while the jointly trained trajectory encoder and meta reward model provide transferable structure for online PbRL on the target task.

\subsection{Optimal Transport}

Optimal transport (OT)~\cite{villani2008optimal,peyre2019computational} provides a principled framework for finding the minimum-cost way to transform one probability distribution into another. A key byproduct of this optimization is an explicit transport plan that establishes soft correspondences between samples from the two distributions.

Given two discrete distributions $\alpha = \sum_{i=1}^{n} a_i \delta_{x_i}$ and $\beta = \sum_{j=1}^{m} b_j \delta_{y_j}$, the Gromov--Wasserstein (GW) distance measures their discrepancy by comparing pairwise relational structures within each distribution rather than absolute sample positions:
\begin{equation}
    \mathcal{D}_{\text{GW}}(\alpha, \beta) =
    \min_{T \in \Pi(a,b)}
    \sum_{i,i',j,j'} L_{ii'jj'} \, T_{ij} \, T_{i'j'}
    \label{eq:gw_distance}
\end{equation}
where $(i, i')$ and $(j, j')$ index sample pairs within $\alpha$ and $\beta$, respectively, and
$L_{ii'jj'} = \bigl| d_{\mathcal{X}}(x_i, x_{i'}) - d_{\mathcal{Y}}(y_j, y_{j'}) \bigr|^2$
quantifies the mismatch between their pairwise intra-domain distances. The transport matrix $T \in \mathbb{R}^{n \times m}$ specifies how much probability mass is assigned from $x_i$ to $y_j$. The feasible set
\begin{equation}
    \Pi(a,b) = \{T \geq 0 \mid T\mathbf{1}_m = a,\; T^\top\mathbf{1}_n = b\}
\end{equation}
enforces the marginal constraints, where $a = [a_1,\ldots,a_n]^\top$ and $b = [b_1,\ldots,b_m]^\top$ denote the source and target probability weights.

Solving Eq.~\eqref{eq:gw_distance} exactly is computationally expensive. In practice, entropic GW solvers based on Sinkhorn iterations~\cite{cuturi2013sinkhorn} are commonly used to obtain an efficient approximation:
\begin{equation}
    T^* =
    \arg\min_{T \in \Pi(a,b)}
    \sum_{i,i',j,j'} L_{ii'jj'} \, T_{ij} \, T_{i'j'}
    - \epsilon H(T)
    \label{eq:gw_sinkhorn}
\end{equation}
where $H(T) = -\sum_{ij} T_{ij} \log T_{ij}$ is the entropy of the transport matrix, and $\epsilon > 0$ controls the smoothness of the solution.

The solved transport matrix $T^*$ establishes soft correspondences between the two sample sets: a large $T^*_{ij}$ indicates that $x_i$ and $y_j$ play analogous structural roles within their respective distributions. These correspondences can be used to propagate preference supervision from labeled to unlabeled trajectory pairs. For example, PEARL~\cite{liu2024pearl} applies GW transport directly in the raw state-action trajectory space to transfer preferences across tasks, while VOTP~\cite{luu2026video} uses the transport over pretrained video foundation model embeddings to propagate labels within a single task. PrefO2O differs from both: it performs GW alignment over a jointly learned, task-invariant preference representation, enabling cross-task and cross-embodiment transfer with robustness to distribution shift.

\begin{figure*}[t]
    \centering
    \includegraphics[width=\linewidth]{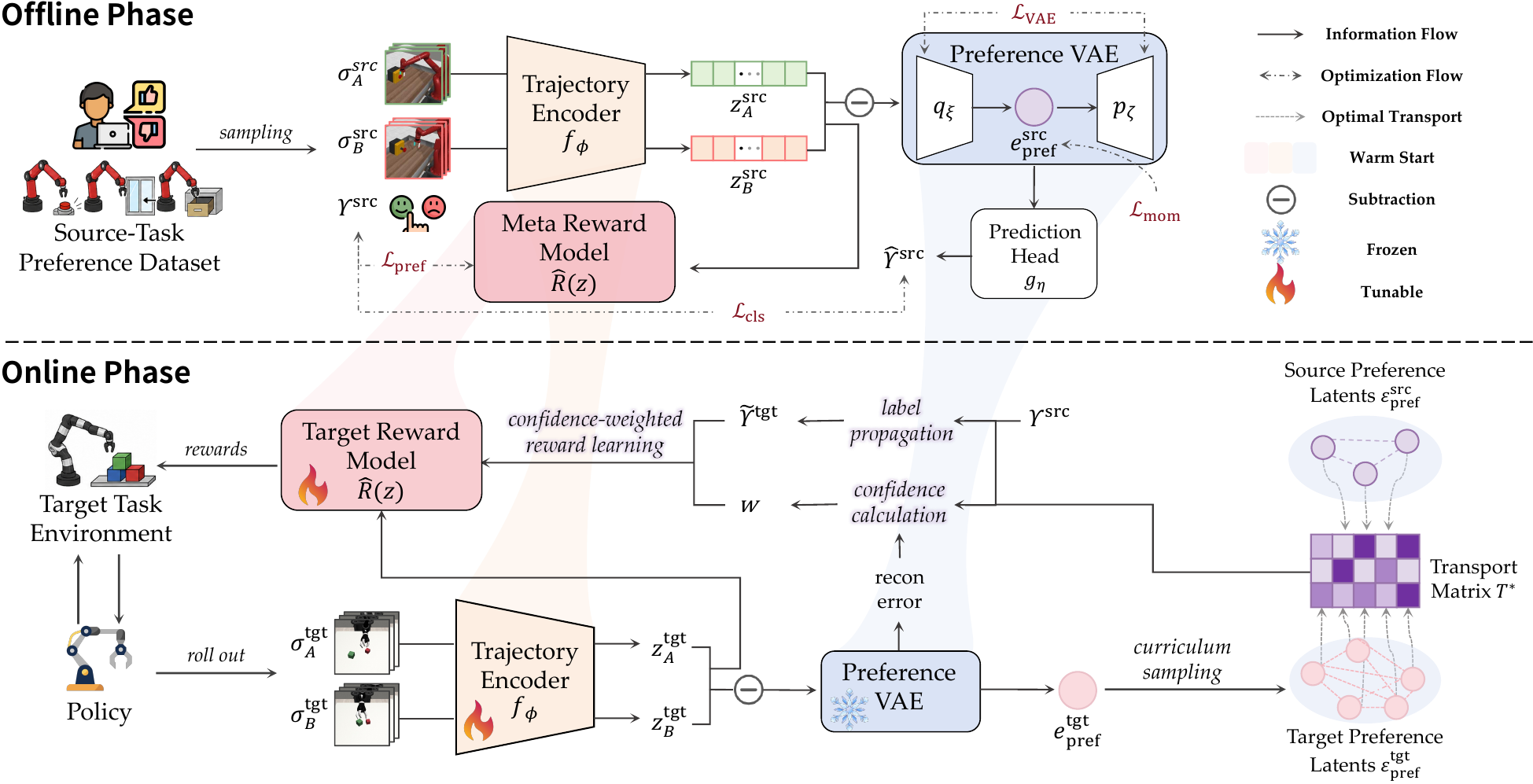}
    \caption{Overview of PrefO2O. The offline phase jointly learns a trajectory encoder, a preference VAE, and a meta reward model from source-task preference data. In the online phase, the pretrained encoder and reward model warm-start target-task PbRL, and the frozen preference VAE serves as a stable reference for OT-based alignment. Source labels are propagated to target trajectory pairs through the transport matrix, producing pseudo-preference labels with OOD-aware confidence weights for online reward learning and policy optimization.}
    \label{fig:framework}
    \vspace{-10pt}
\end{figure*}
\section{Methodology}

\subsection{Problem Setting and Assumptions}

We consider a multi-task PbRL setting in which $K$ source tasks $\{\mathcal{T}^k\}_{k=1}^{K}$ ($K \ge 1$) are available to facilitate learning on an unseen target task $\mathcal{T}^{\text{tgt}}$. Unlike prior works that require all tasks to share the same state-action space, such as the same robot embodiment~\cite{hejna2023few,liu2024pearl}, PrefO2O permits transition dynamics, task objectives, and state-action dimensions to vary across tasks. We assume only that the tasks belong to a related domain family, such as robotic manipulation, in which human preferences exhibit shared underlying structures that are transferable beyond individual task specifications. This aligns with prior observations that semantically related tasks can support transferable representations grounded in shared behavioral structure~\cite{lekkala2026value}.

For each source task $\mathcal{T}^k$, we assume access to an offline preference dataset $\mathcal{D}^k_{\text{pref}} = \{(\sigma^k_{A,i},\,\sigma^k_{B,i},\,\Upsilon^k_i)\}_{i=1}^{M_k}$, where $\Upsilon^k_i \in \{0,1,-1\}$ denotes the preference label for the trajectory pair $(\sigma^k_{A,i},\sigma^k_{B,i})$. For the target task $\mathcal{T}^{\text{tgt}}$, no human preference labels are available. Instead, target trajectories are collected by rolling out the current policy during online RL training, and unlabeled trajectory pairs are sampled from these rollouts for pseudo-preference generation. The objective is to learn a well-performing target policy $\pi^{\text{tgt}}$ without any target-task human preference feedback.

\subsection{Overview}
\label{sec:overview}

PrefO2O leverages offline preference data from source tasks to enable zero-shot PbRL on the target task. The core idea is to perform cross-task preference transfer in a learned task-invariant latent space that absorbs task-specific variations and supports robust transfer under distribution shift. As illustrated in Fig.~\ref{fig:framework}, the framework operates in two stages.

\textit{Offline Latent Preference Learning.} We jointly train three components on source-task preference data: a trajectory encoder that maps raw trajectories from each task into compact behavioral representations, a preference VAE that induces a task-invariant latent preference manifold over pairwise trajectory differences, and a meta reward model that predicts preference-aligned rewards from the learned trajectory representations.

\textit{Online Zero-Shot Preference Transfer.} On the target task, the trajectory encoder and reward model provide a warm start for online PbRL, while the preference VAE is frozen to preserve the learned preference manifold. We perform optimal transport alignment in this shared latent space to establish correspondences between source and target trajectory pairs, generating pseudo-preference labels with OOD-aware confidence weights for online reward learning. The policy is then optimized using the learned reward model within the online PbRL loop.
\subsection{Offline Latent Preference Learning}
\label{sec:offline}

Given the aggregated source-task preference data
$\mathcal{D}^{\text{src}} = \bigcup_{k=1}^{K} \mathcal{D}^k_{\text{pref}}$,
the offline phase learns transferable preference structure by jointly training three components: a trajectory encoder, a preference VAE that induces a task-invariant latent preference space, and a meta reward model that provides a transferable reward prior.

\subsubsection{Trajectory Encoder}

Different tasks may involve different dynamics and state-action dimensions, yet underlying behavioral patterns often share common structure. The trajectory encoder $f_\phi: \sigma \mapsto \mathbb{R}^d$ is designed to capture these transferable behavioral features while accommodating heterogeneous inputs. To this end, we decompose $f_\phi$ into a per-task input adapter and shared representation layers. 

For task $\mathcal{T}^k$ with state-action dimension $d_s^k + d_a^k$, a lightweight adapter $\text{MLP}^k_{\text{embed}}$ projects each state-action pair into a common hidden dimension $d_h$. The adapted step features are then processed by a shared step encoder $\text{MLP}_{\text{step}}$ and aggregated over the trajectory segment by temporal average pooling. A shared projection head maps the pooled representation to a $d$-dimensional trajectory latent:
\begin{equation}
  z^k = f_\phi(\sigma^k)
      = \text{MLP}_{\text{proj}}\!\biggl(
          \frac{1}{H} \sum_{t=1}^{H}
            \text{MLP}_{\text{step}}\bigl(
              \text{MLP}^k_{\text{embed}}(s_t, a_t)
            \bigr)
        \biggr).
  \label{eq:traj_encoder}
\end{equation}

This architecture separates task-specific input adaptation from shared behavioral representation learning: the task-specific adapters accommodate heterogeneous state-action dimensions, while the shared step encoder and temporal pooling produce comparable trajectory-level representations across tasks. This pooling-based design provides a simple and stable summary of preference-relevant behavioral evidence, following prior designs~\cite{hirota2025active,yang2024trajectory}.

For a preference pair, we write $z^k_A = f_\phi(\sigma^k_A)$ and $z^k_B = f_\phi(\sigma^k_B)$ to denote the two encoded trajectories.
\subsubsection{Preference VAE}

Preference learning is inherently comparative; labels reflect the relative quality of two trajectories rather than the absolute merit of either trajectory in isolation. We therefore define the pairwise difference $\Delta z^k = z^k_A - z^k_B \in \mathbb{R}^d$ as the input to the preference encoder. This contrastive representation directly captures the supervisory signal of pairwise preferences while reducing reliance on task-specific absolute trajectory features.

To learn a structured and generalizable latent preference space, we model $\Delta z^k$ with a variational autoencoder whose encoder $q_\xi$ maps each $\Delta z^k$ to a stochastic latent preference vector $e_{\text{pref}} \sim q_\xi(e_{\text{pref}} \mid \Delta z^k)$, and whose decoder $p_\zeta$ reconstructs the input as $\Delta\hat{z} = p_\zeta(e_{\text{pref}})$. Training combines two complementary objectives.

\noindent \textit{a) Latent space regularization.}
The latent space must be smooth, continuous, and distributionally consistent across source tasks to support robust preference transfer during online learning. The standard VAE evidence lower bound encourages a geometrically regular manifold:
\begin{equation}
  \mathcal{L}_{\text{VAE}}
    = \underbrace{\|\Delta z - \Delta\hat{z}\|_2^2}_{\mathcal{L}_{\text{recon}}}
    + \underbrace{D_{\text{KL}}\bigl(
        q_\xi(e_{\text{pref}} \mid \Delta z)
        \,\|\,
        \mathcal{N}(0,I)
      \bigr)}_{\mathcal{L}_{\text{KL}}}
  \label{eq:vae_loss}
\end{equation}

A moment matching constraint further promotes task invariance by penalizing deviations between per-task latent statistics and global source-task statistics:
\begin{equation}
    \mathcal{L}_{\text{mom}} = \sum_{k=1}^{K} \bigl(\|\mu_k - \bar{\mu}\|_2^2 + \|\text{std}_k - \overline{\text{std}}\|_2^2\bigr)
    \label{eq:moment_loss}
\end{equation}
where $\mu_k$ and $\text{std}_k$ denote the empirical mean and standard deviation of the latent preference vectors from source task $k$, and $\bar{\mu}$ and $\overline{\text{std}}$ denote the corresponding global statistics aggregated across all source tasks.

\noindent \textit{b) Preference discrimination.}
A geometrically regular latent space is insufficient if it discards the preference signal. We therefore attach an auxiliary preference head $g_\eta$ to $e_{\text{pref}}$, trained with a cross-entropy loss:
\begin{equation}
  \mathcal{L}_{\text{cls}}
    = \text{CE}\bigl(g_\eta(e_{\text{pref}}),\;\Upsilon\bigr)
  \label{eq:cls_loss}
\end{equation}

This loss explicitly requires the latent space to retain preference-discriminative information, ensuring that the manifold encodes not only behavioral similarity but also preference directionality. Together, these objectives ensure that the latent preference manifold is both preference-discriminative and geometrically regular, while remaining distributionally aligned across tasks.

\subsubsection{Meta Reward Model}

We also train a meta reward model $\hat{R}$ on top of the trajectory latent $z$, using aggregated source-task preference data and the standard preference loss $\mathcal{L}_{\text{pref}}$ in Eq.~\eqref{eq:pref_loss}.

By operating on the learned latent trajectory representation rather than raw state-action inputs, $\hat{R}$ can capture reward patterns that are more readily transferable across tasks. While the preference VAE provides a structured latent space for cross-task alignment, $\hat{R}$ provides a transferable reward prior for online adaptation on the target task. Together, they allow PrefO2O to initialize target-task PbRL with both a preference-structured representation space and a preference-informed reward model.

\subsubsection{Joint Training Objective}

All components are trained end-to-end with a combined objective:
\begin{equation}
  \mathcal{L}_{\text{offline}}
    = \underbrace{
        \mathcal{L}_{\text{pref}} + \lambda_{\text{cls}}\mathcal{L}_{\text{cls}}
      }_{\text{preference modeling}}
    + \underbrace{
        \lambda_{\text{VAE}}\mathcal{L}_{\text{VAE}}
        + \lambda_{\text{mom}}\mathcal{L}_{\text{mom}}
      }_{\text{latent space regularization}}.
  \label{eq:total_loss}
\end{equation}

The trajectory encoder $f_\phi$ receives gradients from all loss terms, as it sits upstream of both the preference VAE and the meta reward model. This ensures that the learned behavioral representations are simultaneously preference-relevant and structurally transferable. 

\subsection{Online Zero-Shot Preference Transfer}
\label{sec:online}

On the target task, PrefO2O initializes online PbRL using the pretrained components and continuously derives pseudo-preference supervision via OT-based alignment in the learned latent preference space. OOD-aware confidence weights then modulate the influence of these pseudo labels on reward learning, which in turn guides downstream policy optimization.

\subsubsection{Warm-Start Initialization}

PrefO2O is agnostic to the choice of online PbRL backbone; in this work, we adopt PEBBLE~\cite{lee2021pebble}, which combines preference-based reward learning with SAC~\cite{haarnoja2018soft} for policy optimization.

Three pretrained components are transferred from the offline phase: the trajectory encoder $f_\phi$, the meta reward model $\hat{R}$, and the preference VAE $(q_\xi, p_\zeta)$. The shared layers of $f_\phi$ inherits its offline parameters directly; the input adapter inherits its offline initialization if the target state-action dimension matches a source task seen during training, and is freshly initialized otherwise.

During online learning, $f_\phi$ is fine-tuned to progressively adapt target trajectories to the learned representation space, whereas the preference VAE is kept frozen to preserve the task-invariant preference manifold. This asymmetry is critical for stable transfer: the frozen VAE provides a fixed geometric reference for preference alignment and OOD estimation, while the adaptive trajectory encoder captures target-specific dynamics without collapsing the latent preference structure.

\subsubsection{Preference Alignment via Optimal Transport}
\label{sec:ot_align}

With the pretrained modules in place, a key challenge in online transfer is generating reliable preference supervision for the target task without any human annotation. PrefO2O addresses this challenge by aligning source and target preference distributions in the learned latent space via optimal transport.

Given a target trajectory pair $(\sigma^{\text{tgt}}_{A},\,\sigma^{\text{tgt}}_{B})$ sampled from online rollouts, we encode it as
\begin{equation}
  \Delta z^{\text{tgt}}
    = z^{\text{tgt}}_{A} - z^{\text{tgt}}_{B},
  \qquad
  e^{\text{tgt}}_{\text{pref}}
    = q_\xi\!\left(\Delta z^{\text{tgt}}\right)
  \label{eq:latent_tgt}
\end{equation}

Source preference latents $\mathcal{E}^{\text{src}}_{\text{pref}} = \{e^{\text{src}}_{\text{pref},i}\}_{i=1}^{M}$ are precomputed during the offline phase and cached. We solve the GW problem in Eq.~\eqref{eq:gw_sinkhorn} over $\mathcal{E}^{\text{src}}_{\text{pref}}$ and the current target batch $\mathcal{E}^{\text{tgt}}_{\text{pref}} = \{e^{\text{tgt}}_{\text{pref},j}\}_{j=1}^{N}$, yielding a transport matrix $T^*$. Each entry $T^*_{ij}$ quantifies the correspondence strength between source pair $i$ and target pair $j$ in the latent preference space.

The pseudo-preference label for the $j$-th target pair is obtained by propagating source labels through the transport plan. Specifically, we compute a normalized soft label distribution
\begin{equation}
  \bar{p}_{j,\Upsilon}
    =
    \frac{
      \sum_{i=1}^{M} T^*_{ij} Y_{i,\Upsilon}
    }{
      \sum_{i=1}^{M} T^*_{ij}
    },
  \qquad
  \tilde{\Upsilon}_j
    =
    \arg\max_{\Upsilon \in \{0,1,-1\}}
    \bar{p}_{j,\Upsilon}
  \label{eq:pseudo_label}
\end{equation}
where $Y \in \{0,1\}^{M \times 3}$ is the one-hot encoding of source preference labels. The pseudo-label therefore reflects a soft vote among source pairs, weighted by their latent-space correspondences to the target pair.

\subsubsection{OOD-Aware Sampling and Reward Learning}
\label{sec:ood_reward}

Obtaining reliable pseudo-preference labels via the OT alignment described above requires managing the distributional discrepancy between source and target domains. Although operating in the learned latent preference space already mitigates much of this gap, we further improve transfer robustness through two strategies: curriculum sampling at the batch level and confidence weighting at the loss level.

\noindent \textit{a) Curriculum sampling.}
Online PbRL accumulates target pairs in the replay buffer faster than any tractable OT solver can process, and uniform random sampling may expose the reward model to many off-manifold pairs whose pseudo-labels are unreliable. 

We therefore construct each OT batch as a curriculum-controlled mixture of in-distribution (ID) and OOD target pairs, classified by preference VAE reconstruction error. The mixture is scheduled to be ID-heavy in early online updates and gradually becomes more OOD-inclusive as training proceeds. Early reward updates therefore rely more heavily on manifold-aligned pairs with more reliable pseudo-labels, whereas later updates incorporate a broader set of target pairs to improve reward coverage. Full details are provided in Appendix~\ref{app:ood_details_hard}.

\noindent \textit{b) Confidence-weighted reward learning.}
Sampling determines which pairs enter the OT batch, but the resulting pseudo-labels can still vary in reliability: even within the ID set, transport plans may be diffuse and reconstructions may remain imperfect. We therefore calibrate each pseudo-label with a per-pair confidence weight combining two complementary signals.

The first measures \emph{transport confidence}. Given the normalized soft label distribution $\bar{p}_{j,\Upsilon}$ in Eq.~\eqref{eq:pseudo_label}, we define
\begin{equation}
  w^{\text{ot}}_j
    =
    \max_{\Upsilon \in \{0,1,-1\}}
    \bar{p}_{j,\Upsilon}
  \label{eq:ot_confidence}
\end{equation}
which is high when the transported source labels agree on a single preference outcome and low when transport mass is spread across labels.

The second measures \emph{manifold proximity}, capturing how close the target encoding lies to the frozen source preference manifold:
\begin{equation}
  w^{\text{vae}}_j
    =
    \exp\!\left(
      -\frac{D_{\text{recon},j}}{\bar{D}_{\text{recon}}}
    \right),
  \qquad
  D_{\text{recon},j}
    =
    \bigl\|
      \Delta z^{\text{tgt}}_j
      -
      \Delta\hat{z}^{\text{tgt}}_j
    \bigr\|_2^2
  \label{eq:vae_confidence}
\end{equation}
where $\Delta\hat{z}^{\text{tgt}}_j = p_\zeta(q_\xi(\Delta z^{\text{tgt}}_j))$, and $\bar{D}_{\text{recon}}$ is the mean reconstruction error over source encodings, used as a normalization baseline.

The combined weight $w_j = w^{\text{ot}}_j \cdot w^{\text{vae}}_j$ is high only when the transport assignment is unambiguous and the target encoding lies close to the source preference manifold. These weights drive reward model adaptation through a weighted Bradley--Terry loss:
\begin{equation}
\begin{split}
  \mathcal{L}^{*}_{\text{pref}}
    = -\sum_{j=1}^{N} w_j \Bigl[
      &\mathbb{I}(\tilde{\Upsilon}_j{=}1)
        \log P_\psi[\sigma^{\text{tgt}}_{A,j} \succ \sigma^{\text{tgt}}_{B,j}]\\
      +\;&\mathbb{I}(\tilde{\Upsilon}_j{=}0)
        \log P_\psi[\sigma^{\text{tgt}}_{B,j} \succ \sigma^{\text{tgt}}_{A,j}]
    \Bigr].
\end{split}
  \label{eq:weighted_bt}
\end{equation}
Gradients are backpropagated through both $\hat{R}$ and $f_\phi$.

To convert the trajectory-level reward model $\hat{R}(z)$ into step-wise signals for RL optimization, we follow~\cite{kim2023preference,zhao2024prefmmt} to apply a sliding window over the most recent $H$ state-action pairs:
\begin{equation}
  r_t =
  \hat{R}\!\left(
    f_\phi\bigl(
      \{(s_{t-H+1},a_{t-H+1}),\ldots,(s_t,a_t)\}
    \bigr)
  \right),
  \label{eq:step_reward}
\end{equation}
where $H$ matches the segment length used during offline training. The policy is then optimized with standard SAC updates, as in PEBBLE. 


\section{Experimental Setup}
\label{sec:experiments}
 
We design our experiments to address the following research questions:
\begin{itemize}[leftmargin=*]
    \item\textbf{RQ1.} Can PrefO2O enable efficient zero-shot PbRL on unseen target tasks compared to existing preference transfer methods and PbRL baselines?
    \item\textbf{RQ2.} How well does PrefO2O generalize across diverse transfer scenarios, including multi-to-one vs.\ one-to-one transfer and intra- vs.\ cross-embodiment settings?
    \item\textbf{RQ3.} What is the contribution of each design component to overall performance?
    \item\textbf{RQ4.} Can the learned latent space capture task-invariant, transferable preference structures?
    \item\textbf{RQ5.} Can PrefO2O be deployed in real-world scenarios?
\end{itemize}

\subsection{Task Environments and Transfer Settings}

\subsubsection{Environments and Target Tasks}

We evaluate PrefO2O on manipulation and locomotion tasks. For manipulation, we use two widely used benchmarks: Meta-World~\cite{yu2020meta} and Robosuite~\cite{zhu2020Robosuite}. From Meta-World, we consider five tasks as the target tasks: \textit{Drawer Close}, \textit{Button Press}, \textit{Door Open}, \textit{Sweep Into}, and \textit{Window Close}. From Robosuite, we consider two tasks: \textit{Stack} and \textit{Pick-Place}. 

For locomotion, we use the DeepMind Control Suite~\cite{tassa2018deepmind} and evaluate on \textit{Walker Walk}. These tasks cover diverse manipulation and locomotion behaviors, enabling us to evaluate preference transfer across different task objectives, dynamics, and embodiment configurations.

\subsubsection{Transfer Settings}
We organize the transfer settings along two axes: \emph{embodiment compatibility} (intra-embodiment vs.\ cross-embodiment) and \emph{source diversity} (multi-to-one vs.\ one-to-one transfer). Multi-to-one transfer serves as the default setting, while one-to-one transfer is included to assess the robustness of PrefO2O when source information is limited to a single task.

\noindent {\textbf{\textit{Intra-embodiment transfer}.}}
In the intra-embodiment setting, source and target tasks are selected from the same benchmark. For Meta-World, each of the five tasks is used as the target task in turn. Given a target task, we randomly select three tasks from the remaining four tasks as source domains with labeled preference data. For example, when \textit{Door Open} is used as the target task, three tasks are sampled from \textit{Drawer Close}, \textit{Button Press}, \textit{Sweep Into}, and \textit{Window Close} as source tasks.

For Robosuite, each task is used as the target task in turn, and the remaining task together with \textit{Lift} are used as source tasks. For example, when \textit{Stack} is the target task, \textit{Pick-Place} and \textit{Lift} are used as source tasks. For the DeepMind Control Suite, we evaluate \textit{Walker Walk} as the target task and use \textit{Walker Stand} and \textit{Walker Run} as source tasks.

\noindent \textbf{\textit{Cross-embodiment transfer.}}
Beyond intra-embodiment transfer, we evaluate whether PrefO2O can transfer preferences across different robot embodiments. We construct cross-embodiment settings between Meta-World and Robosuite. When a Meta-World task is used as the target task, we use Robosuite \textit{Stack} and \textit{Pick-Place} as source tasks. Conversely, when a Robosuite task is used as the target task, we randomly select three Meta-World tasks as source tasks.

\noindent \textbf{\textit{One-to-one transfer.}}
In addition to the default multi-to-one setting, we also evaluate one-to-one transfer on Meta-World. For each target task, we select a single task from the remaining Meta-World tasks as the source task. This setting tests whether PrefO2O can still extract useful transferable preference structure when only one source preference dataset is available. Note that we remove $\mathcal{L}_{\text{mom}}$ in this setting, as no cross-source task statistics are available.

\subsubsection{Offline Preference Data Collection}
We construct offline preference datasets following protocols established in prior preference-learning studies~\cite{kim2023preference,zhao2024prefmmt}. For each source-task dataset, we first collect a mixed-quality trajectory pool by rolling out policies of varying performance levels, then sample trajectory pairs from this pool and obtain 100 human preference annotations per task. Annotators are graduate students and staff with prior exposure to robotic manipulation tasks. 

\subsection{Baselines and Ablation Models}
We compare PrefO2O against baselines from three categories: 

\subsubsection{Cross-Domain Preference Transfer Methods}

\begin{itemize}[leftmargin=*]
    \item \textbf{PrefFew}~\cite{hejna2023few}: PrefFew is a few-shot PbRL method based on model-agnostic meta-learning. For a fair comparison, we train PrefFew using the same source tasks and the same number of source-domain preference labels as PrefO2O. On the target task, PrefFew is additionally provided 100 real human preference labels for adaptation.

    \item \textbf{PEARL}~\cite{liu2024pearl}: Zero-shot preference
transfer via OT alignment in the raw state-action space.  Since
PEARL is formulated for offline transfer and requires matched state-action dimensions across tasks, we implement an enhanced variant that (i) shares PrefO2O's online
backbone and
(ii) zero-pads inputs to the maximum dimensionality for
cross-embodiment compatibility.
\end{itemize}

\subsubsection{Foundation-Model-Based Feedback Methods}
\begin{itemize}[leftmargin=*]
    \item \textbf{RL-VLM-F}~\cite{wang2024rl}: A PbRL method that uses a vision-language model (VLM) to generate synthetic preference labels from visual observations of trajectory rollouts. We use GPT-4o as the backbone VLM and follow the original implementation protocol for prompt construction, image sampling, and reward model training.
\end{itemize}
 
\subsubsection{Oracle Methods}
\begin{itemize}[leftmargin=*]
    \item \textbf{PrefGT}~\cite{lee2021pebble,lee2021b}: A PbRL baseline trained from scratch on the target task using preference labels generated from the ground-truth environment reward as a proxy for human judgment. This serves as an upper bound for PbRL methods with privileged preference supervision.
    \item \textbf{SAC}~\cite{haarnoja2018soft}: Soft Actor-Critic trained directly with ground-truth environment rewards on the target task. This serves as an oracle RL reference representing the performance achievable when perfect reward information is available.
\end{itemize}

To validate the contribution of each design component, we evaluate the following ablations of PrefO2O:
\begin{itemize}[leftmargin=*]
    \item \textbf{w/o Latent Space}: This variant removes the learned latent preference space used for transfer. Specifically, we remove both the trajectory encoder and the preference VAE, train the meta reward model directly on raw trajectory inputs, and perform pseudo-label generation in the raw trajectory space. This ablation evaluates whether transferring preferences in a learned latent space improves robustness over direct alignment in the original trajectory space.
    
    \item \textbf{w/o Preference VAE}: This variant removes the preference VAE and performs OT-based preference alignment directly on trajectory-pair difference embeddings produced by the trajectory encoder. This ablation tests whether the structured latent preference manifold learned by the VAE improves cross-task alignment beyond learned behavioral representations alone.

    \item \textbf{w/o Meta Reward}: This variant does not transfer the pretrained meta reward model to the target task. Instead, the target reward model is randomly initialized and trained only with pseudo-preference labels generated during online learning. This ablation evaluates the benefit of transferring a preference-aligned reward prior from source tasks.
        
    \item \textbf{w/o Joint Training}: This variant trains the trajectory encoder, preference VAE, and meta reward model separately rather than optimizing them with the joint offline objective. This ablation evaluates whether joint optimization is necessary for learning representations that are simultaneously preference-discriminative, geometrically regular, and useful for reward transfer.
    
    \item \textbf{w/o OOD Sampling}: This variant removes the OOD-aware curriculum sampling strategy used to construct OT batches. Instead, target trajectory pairs are sampled randomly from the online replay buffer for OT alignment. 
    
    \item \textbf{w/o OOD Weighting}: This variant removes the confidence weighting mechanism from reward learning. The target reward model is trained under the standard Bradley-Terry loss. 
\end{itemize}

\subsubsection{Evaluation}
We evaluate all methods on all tasks and transfer settings, and conduct ablation studies on \textit{Drawer Close}, \textit{Sweep Into}, \textit{Walker Walk}, and \textit{Stack}. Each experiment is run with five random seeds to ensure statistical robustness. We report success rate for manipulation tasks and episode return for locomotion tasks.

For PrefO2O, applicable ablations, and baselines (PEARL and RL-VLM-F), we additionally report pseudo-label accuracy to assess the quality of generated supervision. Following standard practice~\cite{lee2021pebble}, we use a scripted teacher as the ground-truth oracle, which produces preference labels based on cumulative ground-truth environment rewards. This provides an objective reference for evaluating how accurately each method recovers true preferences without access to environment rewards during training. All hyperparameters are held constant across all experiments.

\section{Results and Analysis}
\label{sec:results}

In this section, we present our experimental results and analysis to answer the research questions posed above.

\begin{figure*}[t]
\centering
\includegraphics[width=\linewidth]{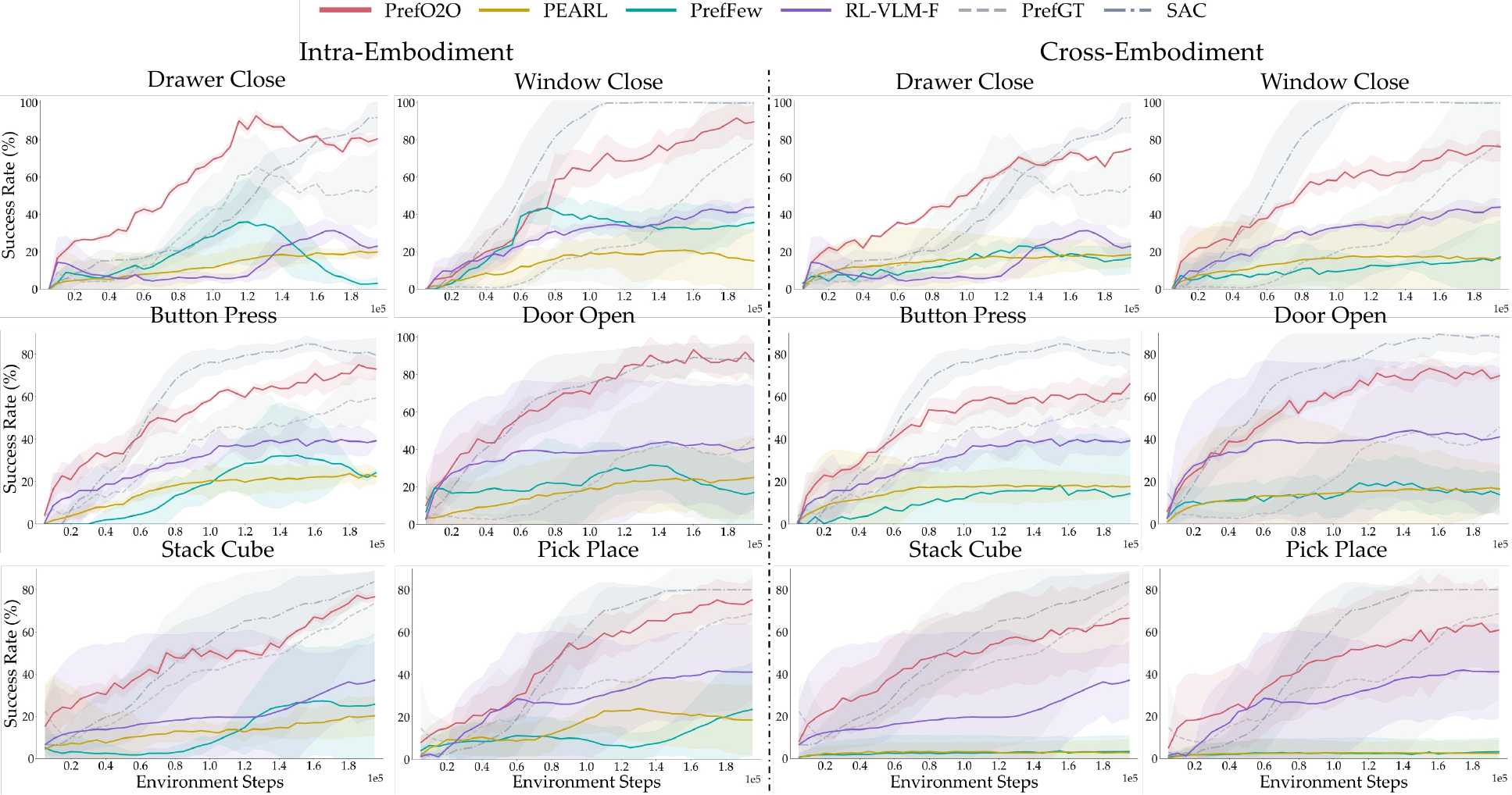}
\caption{Learning curves of all methods on six tasks under two transfer protocols. The left panel shows intra-embodiment transfer, while the right panel shows cross-embodiment transfer. Learning curves for Sweep Into and Walker Walk are provided in Appendix~\ref{appendix:additional_curves}.}
\label{fig:meta_baseline}
\end{figure*}

\subsection{Overall Performance}
\label{sec:overall_results}

\begin{figure}[h]
\centering
\includegraphics[width=\linewidth]{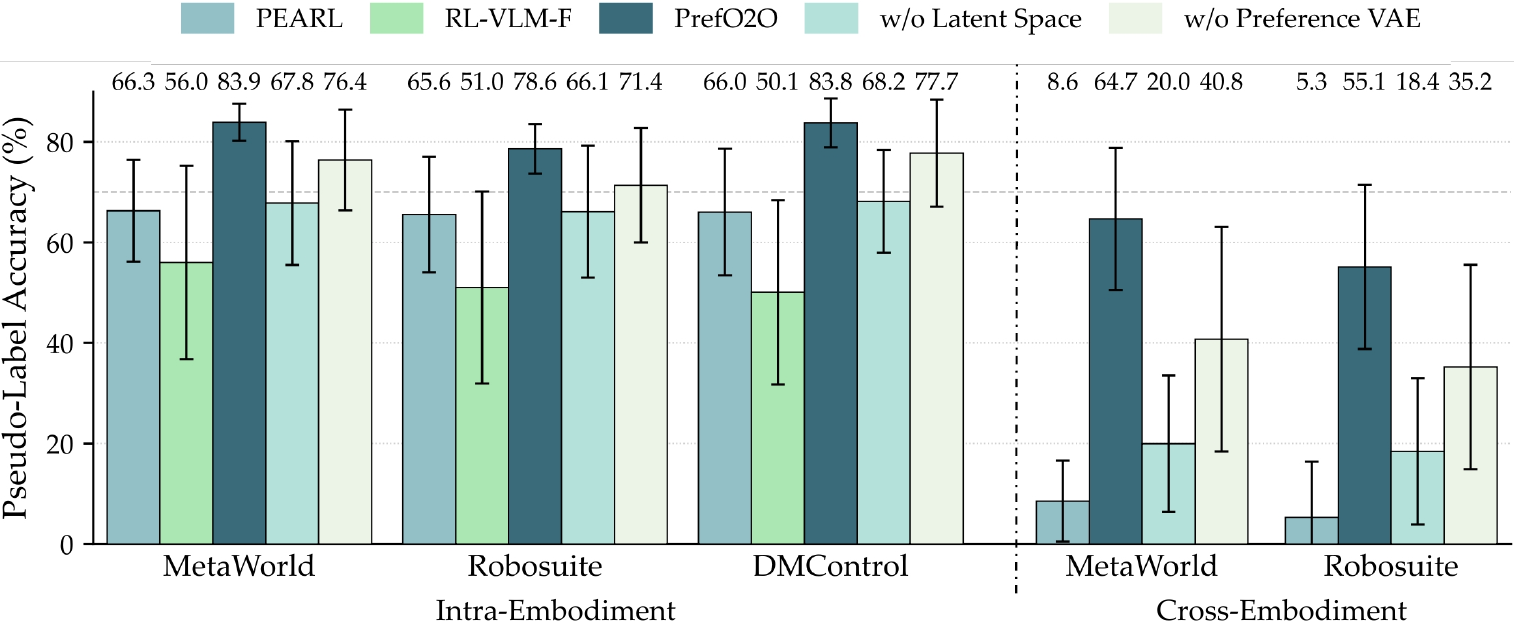}
\caption{Pseudo-label accuracy (\%) across three benchmarks. Bars show the mean over five random seeds; error bars indicate one standard deviation.}
\label{fig:ablation_ot_acc}
\vspace{-5pt}
\end{figure}

Fig.~\ref{fig:meta_baseline} shows the learning curves of all methods across the eight main tasks in MetaWorld and Robosuite under both the intra-embodiment (left) and cross-embodiment (right) protocols. Additional results on the Sweep Into and Walker Walk tasks are reported in Appendix~\ref{appendix:additional_curves}, and pseudo-label accuracies are provided in Fig.~\ref{fig:ablation_ot_acc}.

Across all tasks, PrefO2O consistently surpasses all baselines and matches or even exceeds oracle RL and PbRL methods trained with ground-truth rewards or preference labels, all while achieving fast coverage in early training. This establishes PrefO2O as a strong backbone for zero-shot learning on unseen target tasks: by transferring shared preference structures, it enables rapid adaptation across differing task objectives and environment dynamics, entirely without human feedback on the target. These results directly answer \textbf{RQ1}, confirming that preference structures transfer across heterogeneous tasks effectively enough to support strong zero-shot policy learning on unseen targets.

We first note that PrefFew performs poorly across the board. We attribute this to its reliance on meta-training, which fails to generalize in the \textit{source-limited regime} we consider. Whereas its original formulation assumes roughly ten thousand preference labels spread across ten tasks, our setting provides only one hundred labels per task across three source tasks, mimicking the data scarcity of realistic deployment. This contrast highlights the efficiency of PrefO2O, which transfers effectively from only a handful of source-task preferences and requires zero human labels on the target.

The more informative comparison is against RL-VLM-F and PEARL, over which PrefO2O also yields substantial gains. We attribute these gains to two factors. The first is \textit{the quality of preference supervision} on the target task. As shown in Fig.~\ref{fig:ablation_ot_acc}, PrefO2O consistently attains the highest pseudo-label accuracy across all benchmarks under the intra-embodiment setting, exceeding both RL-VLM-F and PEARL by a clear margin. The advantage widens dramatically under the cross-embodiment setting, where PrefO2O retains most of its accuracy while PEARL and RL-VLM-F both collapse to a small fraction of their intra-embodiment level. RL-VLM-F queries a vanilla VLM to interpret the visual appearance of robot trajectories, so its labels rest on visual plausibility rather than the underlying behavioral preference, rendering them inconsistent and prone to missing task-relevant distinctions. PEARL aligns source and target preferences directly in the raw state-action space, which collapses under embodiment shift because the state-action representation itself changes, leaving no stable basis for alignment. In contrast, PrefO2O performs alignment in a learned task-invariant preference space, yielding robust supervision that degrades only modestly under embodiment change. The high fidelity of these pseudo-labels is precisely what allows PrefO2O to rival oracle methods: when transferred supervision approaches the quality of true preference labels, the absence of target-task feedback ceases to be a performance bottleneck.

The second factor is that PrefO2O transfers preference knowledge not only as label generation but also as \textit{preference-grounded priors} for bootstrapping the target task, such as the pretrained trajectory encoder and meta reward model. This provides the policy with informed initialization rather than learning from scratch, which accounts for the faster coverage observed throughout early training and further explains how PrefO2O can occasionally surpass oracle baselines: even with perfect labels, an oracle still pays the full cost of learning the reward and policy from the ground up, whereas PrefO2O begins from a preference-grounded prior.

\begin{figure*}[t]
\centering
\includegraphics[width=\linewidth]{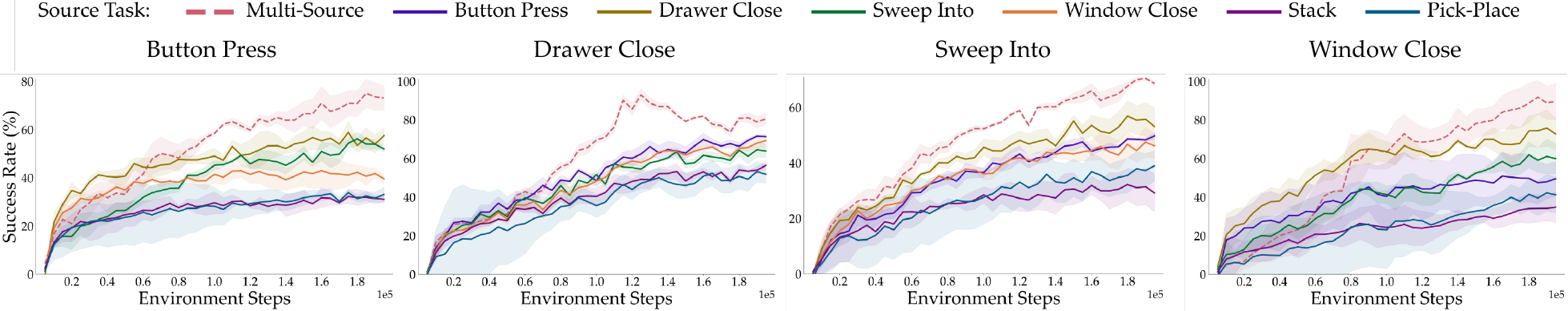}
\caption{Results of one-to-one transfer with intra- and cross-embodiment settings. Different colors indicate different single source tasks used for transfer. The dashed line reports the default multi-to-one performance as a reference.}
\label{fig:one2one}
\end{figure*}

\subsection{Generalization Across Transfer Settings}
\label{sec:generalization_results}

Fig.~\ref{fig:one2one} reports the learning curves for one-to-one transfer on four MetaWorld tasks. Combined with the side-by-side comparison in Fig.~\ref{fig:meta_baseline}, these results show that PrefO2O generalizes well across all four transfer regimes: intra-embodiment, cross-embodiment, multi-source, and single-source.

We first examine robustness to embodiment shift. From the left-right comparison in Fig.~\ref{fig:meta_baseline}, all transfer methods including ours degrade when moving from the intra- to the cross-embodiment setting, as the latter alters the state-action space and transition dynamics, breaking the direct correspondence between source and target trajectories. Crucially, however, PrefO2O degrades only marginally, with a drop of $5$ to $15$ percentage points, whereas the preference-transfer baselines see their relative rankings collapse entirely. 

We attribute this robustness to where preference transfer takes place. Rather than aligning trajectories in the raw state-action space, PrefO2O performs alignment in the learned latent preference space, which is constructed to factor out embodiment-specific information at two levels. The per-task input adapter absorbs dimensional and dynamical heterogeneity before encoding, so the shared backbone operates on a common behavioral abstraction regardless of the underlying embodiment. The moment-matching and preference-discrimination objectives then explicitly shape this space to retain only preference-relevant structure, such as relative motion quality and goal-directed progress, while suppressing absolute, embodiment-dependent features. Because these retained factors remain meaningful across embodiments, the preference correspondences established by optimal transport survive the shift that derails raw-space alignment. The same embodiment-agnostic space lets the shared backbone and meta reward model carry over directly as warm-start priors, so the target policy inherits not only transferable supervision but also a preference-grounded initialization that baselines transferring labels alone cannot provide.

We next examine one-to-one transfer in Fig.~\ref{fig:one2one}, where each solid curve uses a single source task and the dashed curve shows the multi-source reference. The multi-source setting generally achieves the strongest performance, indicating that broader source-task coverage improves transfer. Nevertheless, many single-source curves remain competitive and in some cases approach the multi-source reference, showing that PrefO2O can extract transferable preference structure even from one source task. The performance gap among different single sources further suggests that transfer depends on semantic compatibility between source and target tasks. This supports our view that PrefO2O transfers through pairwise preference contrasts $\Delta z$, which encode relational directions of behavioral improvement rather than task-specific trajectory appearance.

Together, these results answer \textbf{RQ2}, confirming that PrefO2O extracts a task-invariant preference space that supports robust transfer across both embodiment shifts and varying degrees of source-task diversity.

\begin{figure*}[t]
\centering
\includegraphics[width=\linewidth]{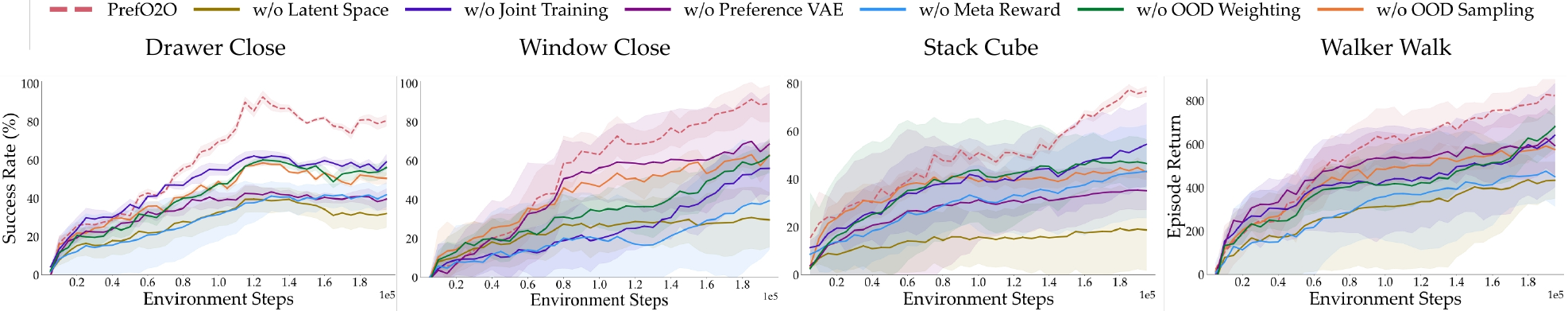}
\caption{Results of ablation study on four target tasks under default intra-embodiment, multi-to-one transfer settings.}
\label{fig:meta_ablation}
\end{figure*}

\begin{figure*}[t]
\centering
\includegraphics[width=\linewidth]{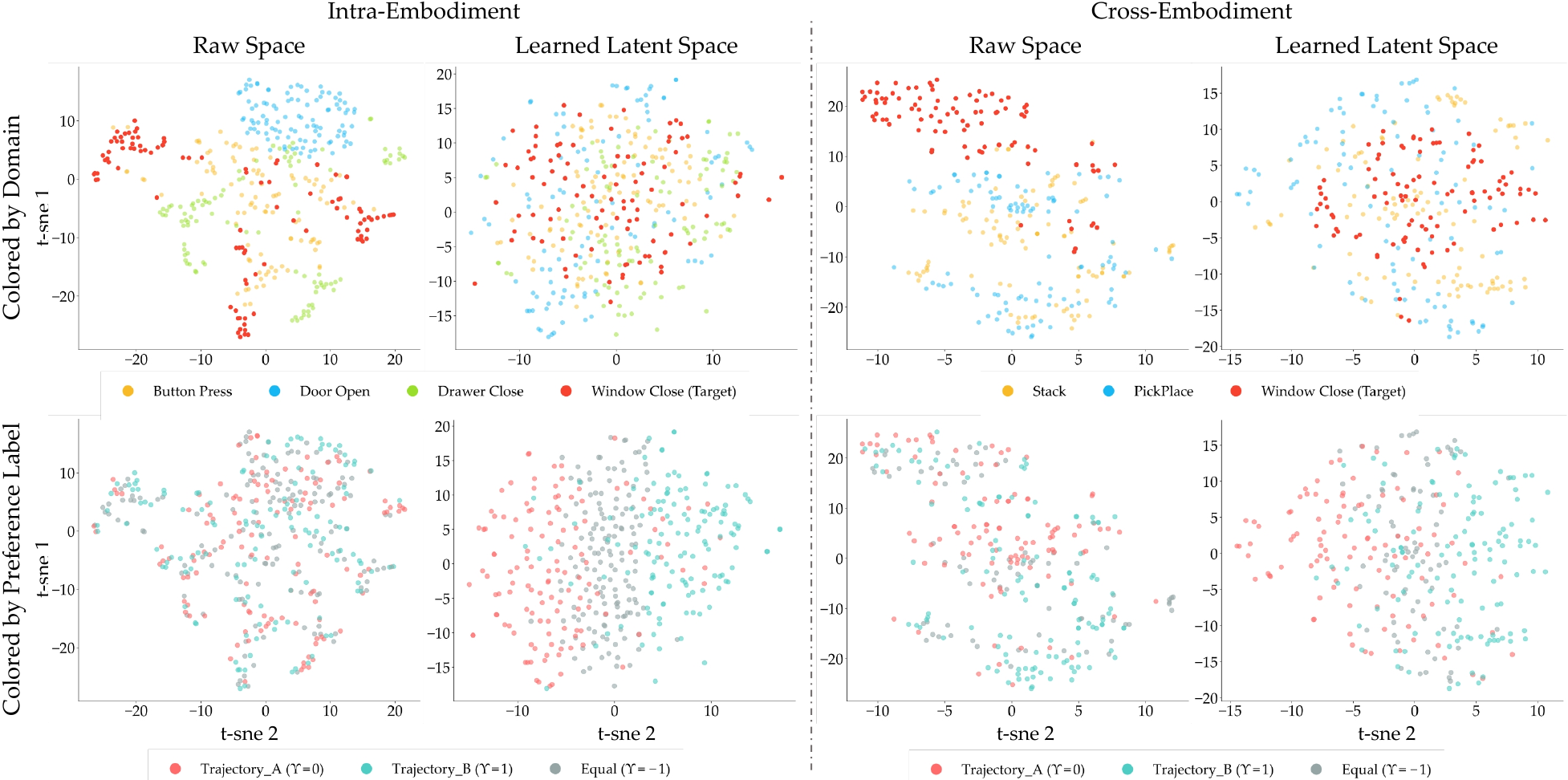}
\caption{t-SNE visualization of preference representations with Window Close as the target task, comparing the raw space against the learned latent preference space under intra-embodiment (left) and cross-embodiment (right) transfer. Each point denotes either a raw trajectory-pair difference or its corresponding latent preference vector $e_{\text{pref}}$. The upper row colors points by task domain and the lower row by preference label $\Upsilon$. The learned latent space exhibits strong cross-task mixing and clear separation across preference labels, indicating a task-invariant yet preference-discriminative structure that the raw space lacks.}
\label{fig:tsne_domain}
\end{figure*}

\subsection{Ablation Study}
\label{sec:ablation_results}
Fig.~\ref{fig:meta_ablation} shows the results of the ablation study, comparing six ablation variants against the full PrefO2O on four target tasks. We can notice that compared to the full version of PrefO2O, every variant degrades in either convergence speed, final performance, or both, and the magnitude of this degradation broadly groups into tiers that mirror the design of our method.

The most consistent degradation comes from removing the latent preference space. Across the four tasks, this variant tends to converge more slowly, reach a lower final performance, and exhibit larger variance, with the gap to the full model being especially visible on Drawer Close and Window Close. This supports our view that the task-invariant latent space is central to effective transfer: aligning preferences directly over raw trajectory representations exposes the alignment to task-specific variation that the latent space is designed to absorb. The two components that shape this space, joint training and the preference VAE, form a closely related group. Their removal is generally less severe but still notable, with both variants learning more slowly early on and tending toward a lower plateau. This suggests that the geometric regularity and preference-discriminative structure imposed during offline learning are what make the latent space transferable rather than merely compact.

A second group concerns the preference-grounded priors carried over for warm-starting. Removing the meta reward model tends to slow early-stage learning more than most other variants, as the policy must then acquire a reward signal from scratch rather than inheriting a preference-aligned prior. In several tasks this variant recovers at later stages once enough pseudo-supervision accumulates, which helps isolate its role: the meta reward model primarily accelerates early adaptation rather than determining the final ceiling. This is consistent with our claim that PrefO2O transfers preference knowledge not only as label generation but also as reusable structure that bootstraps target-task learning.

The third tier comprises the online OOD mechanisms. Removing either OOD sampling or OOD weighting has only a mild effect on Drawer Close and Window Close, but the degradation grows on Stack Cube and Walker Walk. This pattern is informative: in both tasks the source preferences cover the target's decisive behaviors only sparsely, whether the intermediate gait between Walker Stand and Run, or the precise stacking absent from Lift and Pick-Place, so target rollouts more often fall into sparse regions of the preference manifold where pseudo-labels become unreliable. The OOD mechanisms matter most precisely in this regime, deprioritizing and down-weighting off-manifold pairs to keep online pseudo-supervision clean. Their contribution is thus not merely incremental but grows with the difficulty of aligning the target preference signal.

Together, these three tiers explain how PrefO2O achieves both its early-training speed and its final performance, answering \textbf{RQ3}.

\begin{figure*}[t]
\centering
\includegraphics[width=\linewidth]{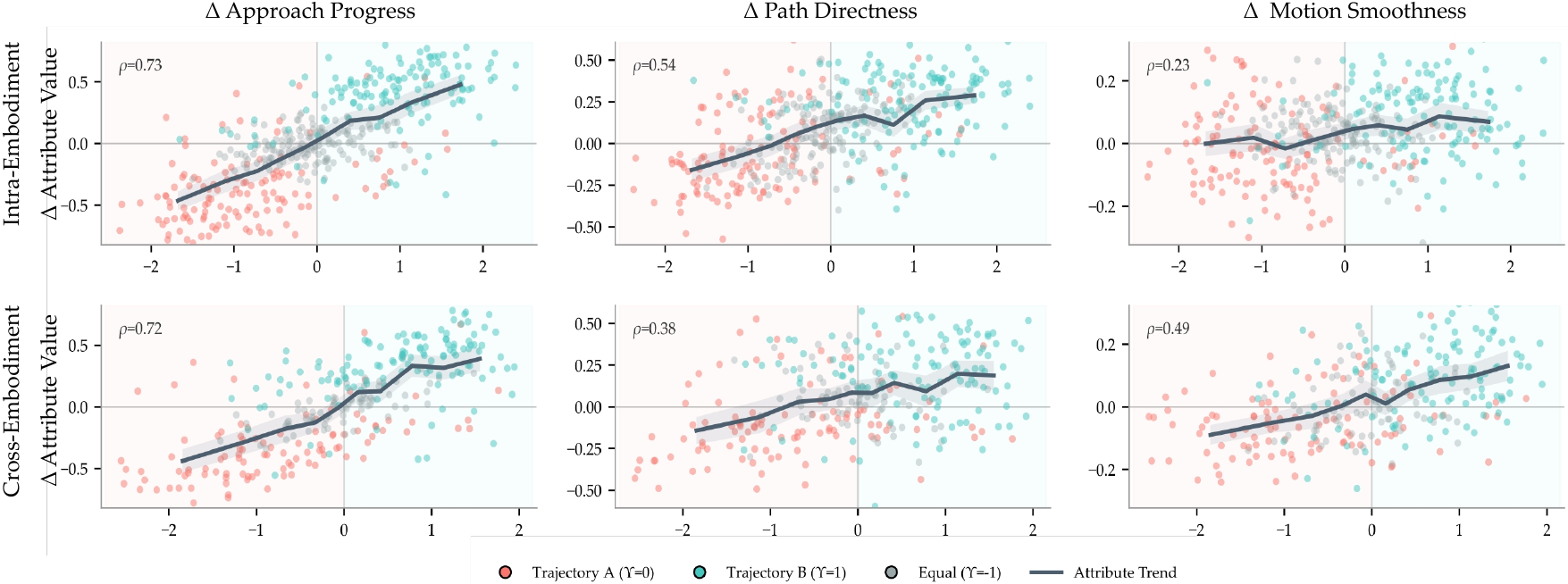}
\caption{
Attribute alignment along the learned preference axis with Window Close as the target task, evaluated under intra-embodiment (top) and cross-embodiment (bottom) transfer.
Each point denotes a latent preference vector $e_{\text{pref}}$, colored by preference label $\Upsilon$.
The horizontal axis is the preference-axis projection score, obtained by projecting $e_{\text{pref}}$ onto the direction between the centroids of samples with $\Upsilon=0$ and $\Upsilon=1$.
The vertical axis reports the attribute value difference for three task-relevant behavioral attributes: approach progress, path directness, and motion smoothness.
The dark curve shows the attribute trend along the preference axis, computed as the binned mean attribute value.
The Spearman correlation coefficient $\rho$ in each panel quantifies the monotonic association between the preference-axis projection score and the corresponding attribute value difference.
The attributes increase consistently along the learned preference axis, indicating that the latent preference space captures a structured behavioral manifold aligned with preference-relevant properties.
}
\label{fig:tsne_attribute}
\end{figure*}

\subsection{Latent Space Analysis}
\label{sec:latent_analysis}

\begin{figure*}[t]
\centering
\includegraphics[width=\linewidth]{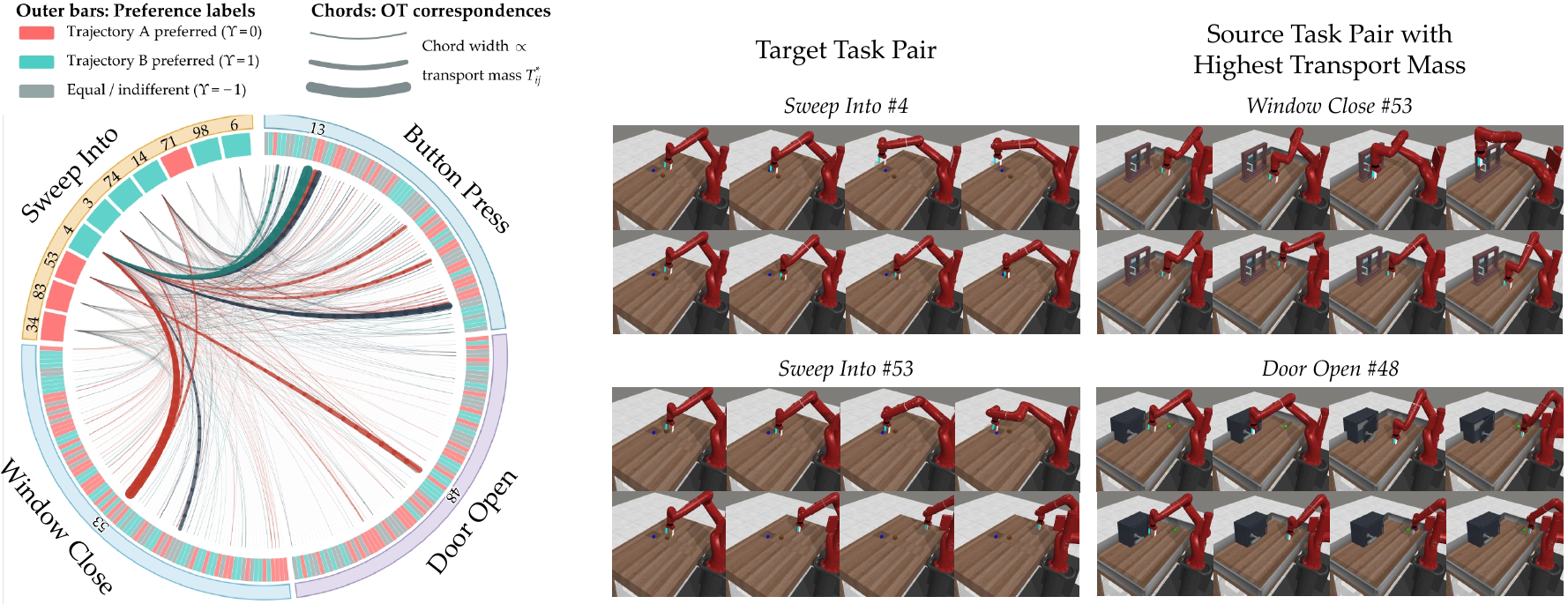}
\caption{Case study of OT-based preference alignment in the learned latent space, with Sweep Into as the target task. \textbf{Left:} a chord diagram where outer bars denote individual preference pairs grouped by source task, colored by preference label $\Upsilon$; each chord connects a target pair to a source pair, with width proportional to the transport mass $T^*_{ij}$. Every target pair draws correspondences from all source tasks, but with markedly different mass, showing that alignment is driven by preference structure rather than task identity. \textbf{Right:} for two representative target pairs (Sweep Into \#4 and \#53), the source pair receiving the highest transport mass, visualized as trajectory rollouts. Despite belonging to entirely different tasks with distinct object interactions, the matched source and target pairs share the same underlying behavioral quality contrast, confirming that the learned latent space aligns preferences along transferable behavioral features rather than task-specific motions.}
\label{fig:sweep_ot}
\end{figure*}

We further analyze whether the latent space learned by PrefO2O captures preference structures that are invariant to task and embodiment, and therefore transferable across tasks. We conduct three complementary analyses using Sweep Into as the target task. Fig.~\ref{fig:tsne_domain} examines the distributional geometry of the learned space, Fig.~\ref{fig:tsne_attribute} analyzes its semantic organization, and Fig.~\ref{fig:sweep_ot} visualizes how the learned space supports OT-based pseudo-label generation.

Fig.~\ref{fig:tsne_domain} compares raw trajectory-pair representations with the learned latent preference representations, colored by task domain in the upper row and by preference label in the lower row. In the raw space, samples are more strongly separated by task domain, and preference labels are more entangled. This suggests that alignment in the raw space can be dominated by task identity rather than preference-relevant similarity. In contrast, the learned latent space shows substantially stronger cross-task mixing while organizing samples more coherently according to preference labels. This pattern appears in both intra- and cross-embodiment settings, indicating that the learned space reduces task- and embodiment-specific variation while preserving preference-discriminative structure.

This raises the question of what the separating preference direction in the learned space actually encodes. We therefore examine whether this direction is semantically meaningful by projecting each latent preference vector onto the learned preference axis.
As shown in Fig.~\ref{fig:tsne_attribute}, the horizontal axis represents the preference-axis projection score, while the vertical axis reports real differences in three interpretable behavioral attributes: Approach Progress, Path Directness, and Motion Smoothness, as defined in Appendix~\ref{app:behavioral_attributes}. Across both intra-embodiment and cross-embodiment settings, the attribute trends increase along the preference axis, indicating that moving in the direction that separates preference labels also corresponds to systematic improvements in behavioral quality. As reflected by the Spearman correlations $\rho$, this trend is strongest for Approach Progress, remains clear for Path Directness, and is weaker but still positive for Motion Smoothness, consistent with a hierarchy in how strongly different behavioral factors contribute to the learned preference structure. Importantly, the trends persist under cross-embodiment transfer, suggesting that the latent space expresses preference through embodiment-independent behavioral contrasts rather than raw kinematic similarity.

Finally, we verify that this semantic structure is what governs transfer in practice. Fig.~\ref{fig:sweep_ot} presents a case study of the OT alignment for Sweep Into target pairs. The chord diagram shows that each target pair draws transport mass from pairs across all source tasks, but concentrated on a small set of dominant matches that consistently share the target's preference label. The trajectory rollouts make the basis of these matches explicit and reveal that different target pairs are aligned through different criteria: Sweep Into \#4 matches Window Close \#53 because both preferences are governed by approaching efficiency, whereas Sweep Into \#53 matches Door Open \#48 because both hinge on whether the manipulator successfully engages the target object rather than on the approach motion itself. That two pairs from the same target task are routed to source pairs in different tasks, according to which behavioral attribute determines their preference, confirms that alignment is driven by transferable behavioral features rather than task identity.

Together, these analyses answer \textbf{RQ4}. The learned latent space is more task- and embodiment-invariant than the raw trajectory space, preserves preference label structure, organizes samples according to interpretable behavioral attributes, and induces meaningful OT correspondences across nominally different tasks. These findings provide mechanistic evidence for the strong zero-shot performance and high pseudo-label accuracy reported above.

\subsection{Real-World Robot Validation}
\label{sec:real_robot}

\begin{figure}[t]
    \centering
    \includegraphics[width=\linewidth]{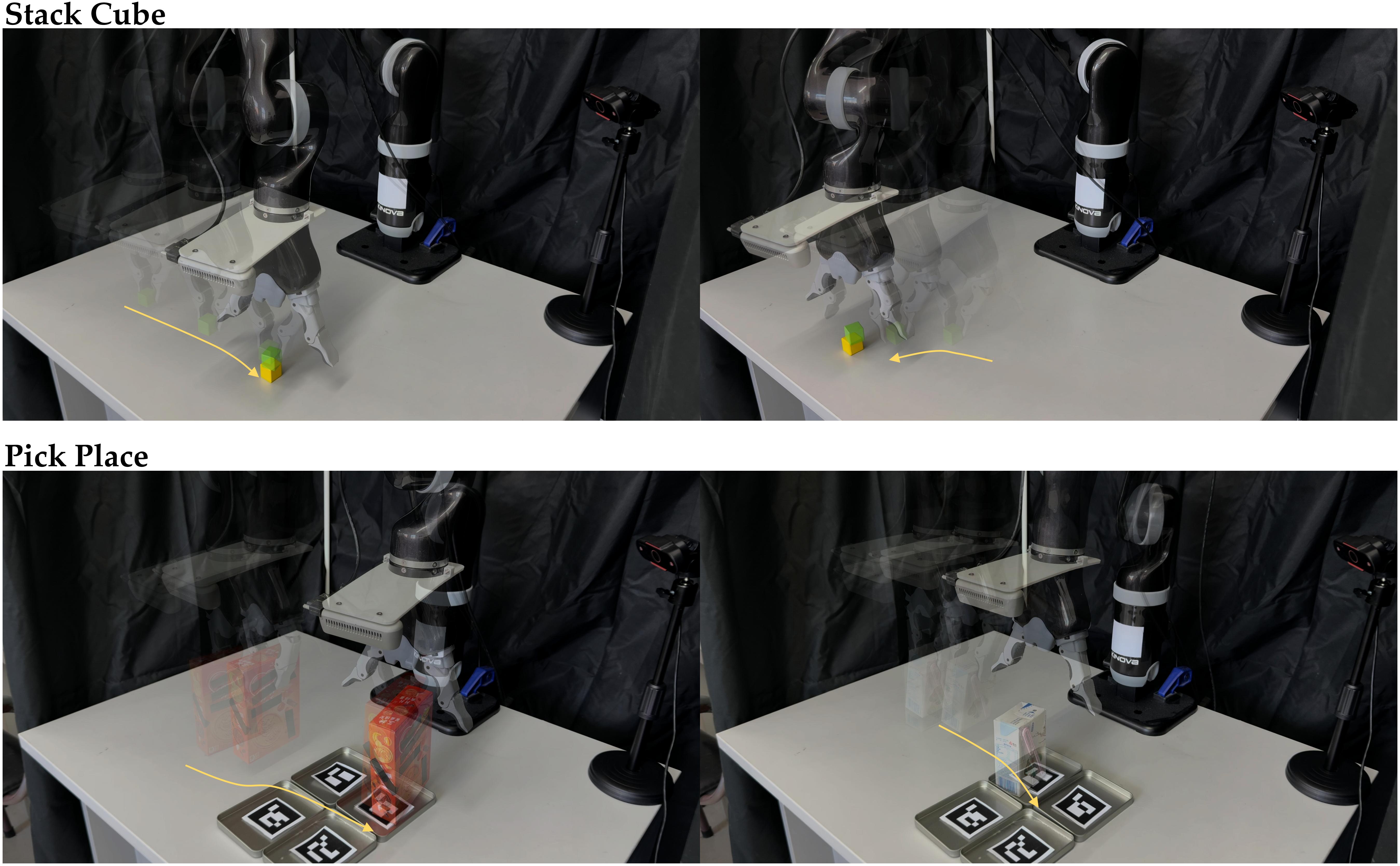}
    \caption{\small PrefO2O successfully completes real-world Cube Stacking and Pick-and-Place tasks on a Kinova Jaco robot.}
    \label{fig:real_demo}
\end{figure}

To validate the practical applicability of PrefO2O, we deploy the trained policies on a physical Kinova Jaco2 arm for two Robosuite tasks~\cite{zhu2020Robosuite}: \textit{Cube Stacking}, where one cube must be placed on top of another, and \textit{Pick-and-Place}, where an object must be grasped and placed into a target container.

Policies are first trained in simulation and then transferred to the physical platform. The
complete hardware configuration, camera setup, sim-to-real settings, and
evaluation protocol are described in Appendix~\ref{app:real_world_setup}. We evaluate each policy over $10$ trials on the physical robot. PrefO2O achieves success rates of $7/10$ on \textit{Cube Stacking} and $6/10$ on \textit{Pick-and-Place}. These results show that policies trained with transferred preference supervision can retain meaningful task performance under real-world perception noise, actuation variability, and contact uncertainty. Fig.~\ref{fig:real_demo} shows representative executions of both tasks.

As a comparative reference, we also deploy PEARL on the same physical platform. Over $10$ trials per task, PEARL achieves success rates of $2/10$ on \textit{Cube Stacking} and $1/10$ on \textit{Pick-and-Place}, trailing PrefO2O on both tasks. The gap is especially pronounced on \textit{Pick-and-Place}, where successful execution requires accurate grasping, transport, and placement. These results answer \textbf{RQ5} by demonstrating that PrefO2O can transfer preference-guided policies to physical robotic settings without collecting target-task human feedback.

\section{Conclusion and Discussion}
\label{sec:conclusion}

We presented PrefO2O, an offline-to-online framework for zero-shot PbRL that transfers preference structures from a small set of labeled source tasks to an unlabeled target task, even across differing objectives, dynamics, and robot embodiments. The central idea is to perform preference transfer not over raw trajectories, but within a learned latent preference space that is jointly trained to be task-invariant and preference-discriminative. 

Across manipulation and locomotion benchmarks under both intra- and cross-embodiment protocols, PrefO2O consistently outperforms preference-transfer baselines and achieves performance competitive with, and in several settings superior to, oracle RL and PbRL methods trained with target-task supervision. It also transfers effectively from as little as a single source task. Our analyses suggest that these gains arise from two complementary mechanisms: the learned latent space provides more reliable source-target preference correspondences than raw trajectory alignment, and the pretrained trajectory encoder and meta reward model provide a strong initialization for target-task learning. More broadly, the results support the view that what transfers across tasks is not the raw trajectory distribution itself, but the relational, behavior-level contrasts that underlie human preference judgments.

PrefO2O has several limitations that suggest directions for future work. First, the approach assumes that source and target tasks belong to a related domain family sharing common behavioral structure; transfer to tasks whose preferences depend on entirely novel behavioral dimensions, absent from all source tasks, remains an open challenge. Second, the optimal transport alignment operates on sampled mini-batches for tractability, and although our curriculum sampling mitigates the resulting variance, scaling alignment to very large source pools is an avenue for further study. Third, our analysis reveals that the latent space spontaneously organizes preferences along interpretable behavioral attributes, which suggests a complementary opportunity: explicitly injecting semantic guidance during training, whether from human-specified behavioral descriptors or from foundation models that articulate the criteria underlying a preference, could yield a more structured and controllable preference manifold, further improving both transfer quality and interpretability.

Overall, PrefO2O represents a step toward general-purpose robots that can learn from human feedback more scalably. Rather than treating each new task as requiring a new round of preference collection, PrefO2O shows that preference knowledge can be accumulated from prior tasks and reused through a task-invariant latent space. This perspective shifts PbRL from task-specific reward learning toward transferable preference learning, where robots bootstrap new behaviors from previously learned human judgments. As robots move toward open-ended real-world deployment, such reuse of preference supervision will be essential for reducing human annotation burden while preserving alignment with human intent. 

\bibliography{main}

@article{kober2013reinforcement,
  title={Reinforcement learning in robotics: A survey},
  author={Kober, Jens and Bagnell, J Andrew and Peters, Jan},
  journal={The International Journal of Robotics Research},
  volume={32},
  number={11},
  pages={1238--1274},
  year={2013},
  publisher={SAGE Publications Sage UK: London, England}
}

@INPROCEEDINGS{wang2024personalization,
  author={Wang, Ruiqi and Zhao, Dezhong and Suh, Dayoon and Yuan, Ziqin and Chen, Guohua and Min, Byung-Cheol},
  booktitle={2025 IEEE International Conference on Robotics and Automation (ICRA)}, 
  title={Personalization in Human-Robot Interaction Through Preference-Based Action Representation Learning}, 
  year={2025},
  volume={},
  number={},
  pages={7377-7384},
  doi={10.1109/ICRA55743.2025.11128756}
}

@article{lekkala2026value,
  title={Value Explicit Pretraining for Learning Transferable Representations},
  author={Lekkala, Kiran and Bao, Henghui and Sontakke, Sumedh A and Biyik, Erdem and Itti, Laurent},
  journal={IEEE robotics and automation letters},
  year={2026},
  publisher={IEEE}
}

@inproceedings{finn2017model,
  title={Model-agnostic meta-learning for fast adaptation of deep networks},
  author={Finn, Chelsea and Abbeel, Pieter and Levine, Sergey},
  booktitle={International conference on machine learning},
  pages={1126--1135},
  year={2017},
  organization={PMLR}
}

@article{wang2025prefclm,
  title={Prefclm: Enhancing preference-based reinforcement learning with crowdsourced large language models},
  author={Wang, Ruiqi and Zhao, Dezhong and Yuan, Ziqin and Obi, Ike and Min, Byung-Cheol},
  journal={IEEE Robotics and Automation Letters},
  year={2025},
  publisher={IEEE}
}

@INPROCEEDINGS{zhao2024prefmmt,
  author={Zhao, Dezhong and Wang, Ruiqi and Suh, Dayoon and Kim, Taehyeon and Yuan, Ziqin and Min, Byung-Cheol and Chen, Guohua},
  booktitle={2025 IEEE/RSJ International Conference on Intelligent Robots and Systems (IROS)}, 
  title={PrefMMT: Modeling Human Preferences in Preference-based Reinforcement Learning with Multimodal Transformers}, 
  year={2025},
  volume={},
  number={},
  pages={18692-18699},
  doi={10.1109/IROS60139.2025.11247027}
}

@article{gupta2022unpacking,
  title={Unpacking reward shaping: Understanding the benefits of reward engineering on sample complexity},
  author={Gupta, Abhishek and Pacchiano, Aldo and Zhai, Yuexiang and Kakade, Sham and Levine, Sergey},
  journal={Advances in Neural Information Processing Systems},
  volume={35},
  pages={15281--15295},
  year={2022}
}

@inproceedings{lee2021pebble,
  title={PEBBLE: Feedback-Efficient Interactive Reinforcement Learning via Relabeling Experience and Unsupervised Pre-training},
  author={Lee, Kimin and Smith, Laura and Abbeel, Pieter},
  booktitle={International Conference on Machine Learning},
  year={2021}
}

@article{lee2021b,
  title={B-Pref: Benchmarking Preference-Based Reinforcement Learning},
  author={Lee, K and Smith, L and Dragan, A and Abbeel, P},
  journal={Neural Information Processing Systems (NeurIPS)},
  year={2021}
}

@inproceedings{hejna2023few,
  title={Few-shot preference learning for human-in-the-loop rl},
  author={Hejna III, Donald Joseph and Sadigh, Dorsa},
  booktitle={Conference on Robot Learning},
  pages={2014--2025},
  year={2023},
  organization={PMLR}
}

@article{huang2025survey,
  title={A survey on hallucination in large language models: Principles, taxonomy, challenges, and open questions},
  author={Huang, Lei and Yu, Weijiang and Ma, Weitao and Zhong, Weihong and Feng, Zhangyin and Wang, Haotian and Chen, Qianglong and Peng, Weihua and Feng, Xiaocheng and Qin, Bing and others},
  journal={ACM Transactions on Information Systems},
  volume={43},
  number={2},
  pages={1--55},
  year={2025},
  publisher={ACM New York, NY}
}

@inproceedings{wang2024rl,
  title={RL-VLM-F: reinforcement learning from vision language foundation model feedback},
  author={Wang, Yufei and Sun, Zhanyi and Zhang, Jesse and Xian, Zhou and Biyik, Erdem and Held, David and Erickson, Zackory},
  booktitle={Proceedings of the 41st International Conference on Machine Learning},
  pages={51484--51501},
  year={2024}
}

@article{cuturi2013sinkhorn,
  title={Sinkhorn distances: Lightspeed computation of optimal transport},
  author={Cuturi, Marco},
  journal={Advances in neural information processing systems},
  volume={26},
  year={2013}
}

@article{peyre2019computational,
  title={Computational optimal transport: With applications to data science},
  author={Peyr{\'e}, Gabriel and Cuturi, Marco and others},
  journal={Foundations and Trends{\textregistered} in Machine Learning},
  volume={11},
  number={5-6},
  pages={355--607},
  year={2019},
  publisher={Now Publishers, Inc.}
}

@book{villani2008optimal,
  title={Optimal transport: old and new},
  author={Villani, C{\'e}dric and others},
  volume={338},
  year={2008},
  publisher={Springer}
}

@inproceedings{yu2020meta,
  title={Meta-world: A benchmark and evaluation for multi-task and meta reinforcement learning},
  author={Yu, Tianhe and Quillen, Deirdre and He, Zhanpeng and Julian, Ryan and Hausman, Karol and Finn, Chelsea and Levine, Sergey},
  booktitle={Conference on robot learning},
  pages={1094--1100},
  year={2020},
  organization={PMLR}
}

@article{zhu2020robosuite,
  title={robosuite: A modular simulation framework and benchmark for robot learning},
  author={Zhu, Yuke and Wong, Josiah and Mandlekar, Ajay and Mart{\'\i}n-Mart{\'\i}n, Roberto and Joshi, Abhishek and Nasiriany, Soroush and Zhu, Yifeng},
  journal={arXiv preprint arXiv:2009.12293},
  year={2020}
}

@article{tassa2018deepmind,
  title={Deepmind control suite},
  author={Tassa, Yuval and Doron, Yotam and Muldal, Alistair and Erez, Tom and Li, Yazhe and Casas, Diego de Las and Budden, David and Abdolmaleki, Abbas and Merel, Josh and Lefrancq, Andrew and others},
  journal={arXiv preprint arXiv:1801.00690},
  year={2018}
}

@article{bradley1952rank,
  title={Rank analysis of incomplete block designs: I. The method of paired comparisons},
  author={Bradley, Ralph Allan and Terry, Milton E},
  journal={Biometrika},
  volume={39},
  number={3/4},
  pages={324--345},
  year={1952},
  publisher={JSTOR}
}

@inproceedings{haarnoja2018soft,
  title={Soft actor-critic: Off-policy maximum entropy deep reinforcement learning with a stochastic actor},
  author={Haarnoja, Tuomas and Zhou, Aurick and Abbeel, Pieter and Levine, Sergey},
  booktitle={International conference on machine learning},
  pages={1861--1870},
  year={2018},
  organization={Pmlr}
}

@article{luu2026video,
  title={Video-Based Optimal Transport for Feedback-Efficient Offline Preference-Based Reinforcement Learning},
  author={Luu, Tung M and Kim, Hwanhee and Lee, Younghwan and Yoo, Chang D},
  journal={arXiv preprint arXiv:2606.16856},
  year={2026}
}

@article{wang2026preference,
  title={Preference-based Reinforcement Learning for Robotics: Advancements and Open Challenges},
  author={Wang, Ruiqi},
  journal={Authorea Preprints},
  year={2026},
  publisher={Authorea}
}

@inproceedings{liu2025vlp,
  title={Vlp: Vision-language preference learning for embodied manipulation},
  author={Liu, Runze and Bai, Chenjia and Lyu, Jiafei and Sun, Shengjie and Du, Yali and Li, Xiu},
  booktitle={Proceedings of the 2025 Conference on Empirical Methods in Natural Language Processing},
  pages={28416--28432},
  year={2025}
}

@inproceedings{
wang2025primt,
title={{PRIMT}: Preference-based Reinforcement Learning with Multimodal Feedback and Trajectory Synthesis from Foundation Models},
author={Ruiqi Wang and Dezhong Zhao and Ziqin Yuan and Tianyu Shao and Guohua Chen and Dominic Kao and Sungeun Hong and Byung-Cheol Min},
booktitle={The Thirty-ninth Annual Conference on Neural Information Processing Systems},
year={2025}
}

@article{christiano2017deep,
  title={Deep reinforcement learning from human preferences},
  author={Christiano, Paul F and Leike, Jan and Brown, Tom and Martic, Miljan and Legg, Shane and Amodei, Dario},
  journal={Advances in neural information processing systems},
  volume={30},
  year={2017}
}

@inproceedings{liu2024pearl,
  title={PEARL: Zero-shot Cross-task Preference Alignment and Robust Reward Learning for Robotic Manipulation},
  author={Liu, Runze and Du, Yali and Bai, Fengshuo and Lyu, Jiafei and Li, Xiu},
  booktitle={International Conference on Machine Learning},
  pages={30946--30964},
  year={2024},
  organization={PMLR}
}

@inproceedings{kim2023preference,
  title={PREFERENCE TRANSFORMER: MODELING HUMAN PREFERENCES USING TRANSFORMERS FOR RL},
  author={Kim, Changyeon and Park, Jongjin and Shin, Jinwoo and Lee, Honglak and Abbeel, Pieter and Lee, Kimin},
  booktitle={11th International Conference on Learning Representations, ICLR 2023},
  year={2023},
  organization={International Conference on Learning Representations, ICLR}
}

@inproceedings{peng2024pragmatic,
  title={Pragmatic Feature Preferences: Learning Reward-Relevant Preferences from Human Input},
  author={Peng, Andi and Sun, Yuying and Shu, Tianmin and Abel, David},
  booktitle={International Conference on Machine Learning},
  pages={40258--40270},
  year={2024},
  organization={PMLR}
}

@inproceedings{myers2023active,
  title={Active Reward Learning from Online Preferences},
  author={Myers, Vivek and B{\i}y{\i}k, Erdem and Sadigh, Dorsa},
  booktitle={2023 IEEE International Conference on Robotics and Automation (ICRA)},
  pages={7511--7518},
  year={2023},
  organization={IEEE}
}

@article{torne2026freeform,
  title={Freeform Preference Learning for Robotic Manipulation},
  author={Torne, Marcel and Mahajan, Anubha and Bhat, Abhijnya and Finn, Chelsea},
  journal={arXiv preprint arXiv:2606.32027},
  year={2026}
}

@inproceedings{wilde2020active,
  title={Active preference learning using maximum regret},
  author={Wilde, Nils and Kuli{\'c}, Dana and Smith, Stephen L},
  booktitle={2020 IEEE/RSJ International Conference on Intelligent Robots and Systems (IROS)},
  pages={10952--10959},
  year={2020},
  organization={IEEE}
}

@inproceedings{brown2019extrapolating,
  title={Extrapolating beyond suboptimal demonstrations via inverse reinforcement learning from observations},
  author={Brown, Daniel and Goo, Wonjoon and Nagarajan, Prabhat and Niekum, Scott},
  booktitle={International conference on machine learning},
  pages={783--792},
  year={2019},
  organization={PMLR}
}

@inproceedings{hwang2026causally,
  title={Causally Robust Reward Learning from Reason-Augmented Preference Feedback},
  author={Hwang, Minjune and Korkmaz, Yigit and Seita, Daniel and Biyik, Erdem},
  booktitle={14th International Conference on Learning Representations},
  year={2026},
  organization={International Conference on Learning Representations, ICLR}
}

@article{baraka2026human,
  title={Human-interactive robot learning: Definition, challenges, and recommendations},
  author={Baraka, Kim and Idrees, Ifrah and Kessler Faulkner, Taylor and Biyik, Erdem and Booth, Serena and Chetouani, Mohamed and Grollman, Daniel H and Saran, Akanksha and Senft, Emmanuel and Tulli, Silvia and others},
  journal={ACM Transactions on Human-Robot Interaction},
  volume={15},
  number={2},
  pages={1--31},
  year={2026},
  publisher={ACM New York, NY}
}

@article{biyik2024active,
  title={Active preference-based Gaussian process regression for reward learning and optimization},
  author={B{\i}y{\i}k, Erdem and Huynh, Nicolas and Kochenderfer, Mykel J and Sadigh, Dorsa},
  journal={The International Journal of Robotics Research},
  volume={43},
  number={5},
  pages={665--684},
  year={2024},
  publisher={SAGE Publications Sage UK: London, England}
}

@article{yuan2026prefmoe,
  title={PrefMoE: Robust Preference Modeling with Mixture-of-Experts Reward Learning},
  author={Yuan, Ziqin and Wang, Ruiqi and Zhao, Dezhong and Yang, Baijian and Min, Byung-Cheol},
  journal={arXiv preprint arXiv:2605.00384},
  year={2026}
}

@article{hirota2025active,
  title={Active reward learning and iterative trajectory improvement from comparative language feedback},
  author={Hirota, Eisuke and Yang, Zhaojing and Hiranaka, Ayano and Jun, Miru and Tien, Jeremy and Russell, Stuart J and Dragan, Anca and B{\i}y{\i}k, Erdem},
  journal={The International Journal of Robotics Research},
  pages={02783649251394742},
  year={2025},
  publisher={SAGE Publications Sage UK: London, England}
}

@inproceedings{yang2024trajectory,
  title={Trajectory Improvement and Reward Learning from Comparative Language Feedback},
  author={Yang, Zhaojing and Jun, Miru and Tien, Jeremy and Russell, Stuart and Dragan, Anca and Biyik, Erdem},
  booktitle={8th Annual Conference on Robot Learning},
  year={2024}
}
\bibliographystyle{IEEEtran}

\appendices

\section{Details of OOD-Aware Curriculum Sampling}
\label{app:ood_details_hard}

\subsection{VAE Reconstruction-Error-Based OOD Detection}
\label{app:recon_ood}

The frozen preference VAE, trained exclusively on source encodings,
provides a natural measure of how far each target pair lies from the
source manifold.  Given a target preference encoding $\Delta z^{\text{tgt}}_j$
and its reconstruction $\Delta\hat{z}^{\text{tgt}}_j = p_\zeta(q_\xi(\Delta z^{\text{tgt}}_j))$,
the OOD score is the per-sample reconstruction error:
\begin{equation}
  D_{\text{recon},j} = \bigl\|\Delta z^{\text{tgt}}_j - \Delta\hat{z}^{\text{tgt}}_j\bigr\|_2^2.
  \label{eq:recon_dist}
\end{equation}
Unlike pairwise distance metrics that require explicit comparison
against the source set, this score is computed per-sample in a single
forward pass through the frozen VAE, incurring negligible overhead
during online learning.

For calibration, we precompute a source-side baseline distribution:
\begin{equation}
  \mathcal{D}_{\text{base}}
    = \bigl\{ D_{\text{recon}}(e^{\text{src}}_{\text{pref},i})
              : e^{\text{src}}_{\text{pref},i} \in \mathcal{E}^{\text{src}}_{\text{pref}} \bigr\},
\end{equation}
which characterizes the intrinsic reconstruction tightness of the
learned manifold and serves as a reference for distinguishing
on-manifold target pairs from off-manifold ones.

\subsection{Curriculum Schedule and Adaptive Threshold}
\label{app:threshold}

We adopt a cosine schedule for the OOD mixing weight $\alpha(t) \in [\alpha_{\min}, 1]$:
\begin{equation}
  \alpha(t) = \alpha_{\min}
              + (1 - \alpha_{\min}) \cdot
                \tfrac{1}{2}\!\left(1 - \cos\!\left(\pi t / T\right)\right),
  \label{eq:curriculum}
\end{equation}
where $t$ is the current online step, $T$ is the total online budget,
and $\alpha_{\min} = 0.2$ in all experiments.  The schedule satisfies
$\alpha(0) = \alpha_{\min}$ and $\alpha(T) = 1$: training starts in an
ID-heavy regime (only $20\%$ of each batch drawn from OOD) and
gradually shifts toward OOD-heavy batches as $f_\phi$ adapts.  This
ordering reflects two principles.  First, the reward model benefits
from reliable supervision during early bootstrapping, which on-manifold
pairs provide.  Second, as the encoder progressively pulls target
encodings toward the source manifold, the OOD pool both shrinks and
becomes more meaningful (i.e., genuinely novel rather than simply
under-encoded), making it productive to integrate.

The ID/OOD classification threshold is coupled directly to $\alpha(t)$:
\begin{equation}
  \tau(t) = \mathrm{Quantile}\bigl(\mathcal{D}_{\text{base}},\ 1 - \alpha(t)\bigr),
  \label{eq:threshold}
\end{equation}
classifying target pair $j$ as ID if $D_{\text{recon},j} < \tau(t)$
and as OOD otherwise.  Early in training, $\alpha(t)$ is small, so
$\tau(t)$ is high (a generous quantile) and the ID set is broad;
later, $\tau(t)$ tightens, restricting the ID set to pairs that
genuinely lie close to the manifold.  Coupling $\tau$ to $\alpha$
removes the need for a separately tuned threshold schedule and ensures
the classification boundary co-evolves with the mixing ratio.

\subsection{Diversity-Aware Sampling Within Each Set}
\label{app:sampling_weights}

Given a batch budget $B$, the curriculum allocates
$n_{\text{OOD}} = \lfloor B \cdot \alpha(t) \rfloor$ slots to OOD pairs
and $n_{\text{ID}} = B - n_{\text{OOD}}$ slots to ID pairs.  We draw
ID samples first to establish anchors in the embedding space, then
draw OOD samples whose diversity is measured relative to those anchors.

\paragraph{ID sampling.}
Within the ID set, we sample with weights that prioritize pairs
closest to the source manifold:
\begin{equation}
  w_{\text{ID}}(i)
    \propto \exp\!\left(-\frac{D_{\text{recon},i}}{\bar{D}_{\text{recon}}}\right),
  \label{eq:id_weight}
\end{equation}
where $\bar{D}_{\text{recon}}$ is the mean reconstruction error over
the current ID candidate pool.  This emphasizes pairs whose OT
pseudo-labels are most reliable, providing the cleanest training
signal during early adaptation.

\paragraph{OOD sampling.}
Within the OOD set, we balance two desiderata: pairs should be
informatively distant from the source manifold, and they should
collectively cover diverse regions of the embedding space rather than
clustering in one off-manifold direction.  We combine these into a
single sampling weight:
\begin{equation}
  w_{\text{OOD}}(i)
    \propto \frac{D_{\text{recon},i}}{D_{\text{recon},\max}}
            \cdot
            \min_{j \in \mathcal{S}} \bigl\| z_i - z_j \bigr\|_2,
  \label{eq:ood_weight}
\end{equation}
where $D_{\text{recon},\max}$ normalizes the novelty term to $[0,1]$,
and $\mathcal{S}$ is the set of ID anchors already drawn in the
current batch.  The first factor rewards genuine novelty; the second
penalizes redundancy by pushing OOD samples away from existing
anchors.  Their product produces batches that span the embedding
space both within and beyond the source manifold.

\subsection{Procedure}
\label{app:algorithm}

The full sampling procedure is summarized in
Algorithm~\ref{alg:curriculum}.  Each invocation computes
reconstruction errors for all pairs in the replay buffer, classifies
them under the current threshold $\tau(t)$, and draws a mixed batch
according to $\alpha(t)$.  All operations are $O(|\mathcal{B}|)$
in the buffer size apart from the diversity term in OOD sampling,
which is $O(n_{\text{ID}})$ per OOD candidate; in practice this
incurs no measurable overhead at the batch sizes used in our
experiments.

\begin{algorithm}[t]
\caption{Curriculum Sampling for Online Reward Learning}
\label{alg:curriculum}
\begin{algorithmic}[1]
\Require Replay buffer $\mathcal{B}$, frozen VAE $(q_\xi, p_\zeta)$,
         source baseline $\mathcal{D}_{\text{base}}$, current step $t$,
         total steps $T$, batch budget $B$
\State Compute $\Delta z_i$, $D_{\text{recon},i}$ for each pair in $\mathcal{B}$
       \hfill $\triangleright$ Eq.~\eqref{eq:recon_dist}
\State Evaluate $\alpha(t)$ and $\tau(t) \leftarrow
       \mathrm{Quantile}(\mathcal{D}_{\text{base}},\ 1 - \alpha(t))$
       \hfill $\triangleright$ Eq.~\eqref{eq:curriculum},\eqref{eq:threshold}
\State Partition $\mathcal{B}$:
       $\mathcal{I} \leftarrow \{i : D_{\text{recon},i} < \tau(t)\}$,\quad
       $\mathcal{O} \leftarrow \{i : D_{\text{recon},i} \geq \tau(t)\}$
\State $n_{\text{OOD}} \leftarrow \lfloor B \cdot \alpha(t) \rfloor$,\quad
       $n_{\text{ID}} \leftarrow B - n_{\text{OOD}}$
\State Sample $n_{\text{ID}}$ pairs from $\mathcal{I}$ with weights
       $w_{\text{ID}}$ \hfill $\triangleright$ Eq.~\eqref{eq:id_weight}
\State $\mathcal{S} \leftarrow$ embeddings of the sampled ID pairs
\State Sample $n_{\text{OOD}}$ pairs from $\mathcal{O}$ with weights
       $w_{\text{OOD}}$ \hfill $\triangleright$ Eq.~\eqref{eq:ood_weight}
\State \Return combined batch of $B$ pairs
\end{algorithmic}
\end{algorithm}

\section{Additional Baseline Learning Curves}
\label{appendix:additional_curves}
\begin{figure}[t]
    \centering
    \includegraphics[width=\linewidth]{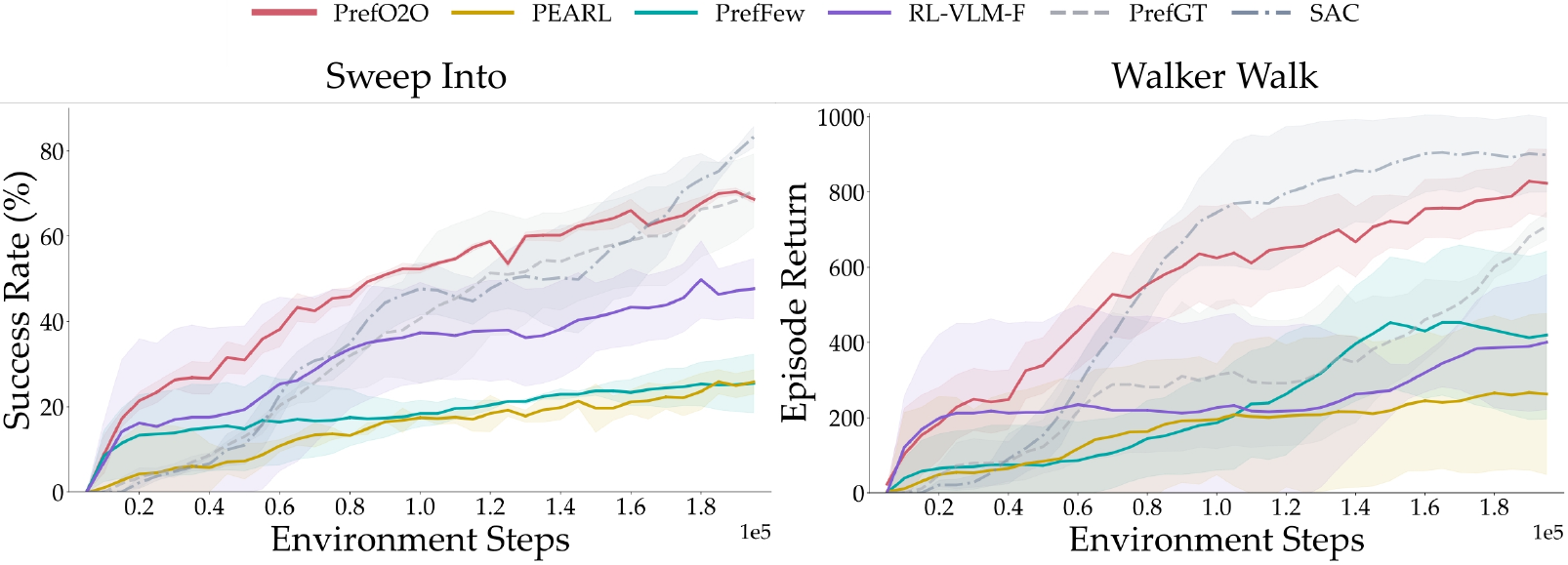}
    \caption{Learning curves on Sweep Into from MetaWorld and Walker Walk from DeepMind Control Suite under the same-embodiment transfer protocol. }
    \label{fig:appendix_baseline}
\end{figure}

This appendix provides the learning curves for Sweep Into from MetaWorld and Walker Walk from DeepMind Control Suite, which were deferred from Fig.~\ref{fig:meta_baseline}. The experimental protocol is identical to that of the main paper. 

Fig.~\ref{fig:appendix_baseline} shows the results. The trends mirror those reported in Fig.~\ref{fig:meta_baseline} of the main paper. Both tasks confirm that the warm-started initialization and latent-space preference alignment of our method generalize beyond the tasks shown in the main panel, including the locomotion domain where dynamics differ substantially from manipulation.

\section{Behavioral Attribute Contrast}
\label{app:behavioral_attributes}

This appendix defines the three behavioral attributes used in Fig.~\ref{fig:tsne_attribute}. Each attribute
assigns a scalar score $\sigma_k(\tau)$ to a trajectory segment
$\tau = \{(\mathbf{s}_t, \mathbf{a}_t)\}_{t=1}^{H}$, where
$\mathbf{x}_t$ denotes the end-effector position and $\mathbf{g}_t$
the position of the task-relevant object at step $t$, with
$d_t = \lVert \mathbf{x}_t - \mathbf{g}_t \rVert$.

\paragraph{Approach Progress.}
How much closer the end-effector gets to the object over the segment:
\begin{equation}
\sigma_{\mathrm{AP}}(\tau) = d_1 - d_H .
\end{equation}
A large positive value means the robot steadily approaches the object;
a value near zero or negative means it stalls or moves away.

\paragraph{Path Directness.}
The ratio between the straight-line displacement and the actual path
length of the end-effector:
\begin{equation}
\sigma_{\mathrm{PD}}(\tau) =
\frac{\lVert \mathbf{x}_H - \mathbf{x}_1 \rVert}
     {\sum_{t=1}^{H-1} \lVert \mathbf{x}_{t+1} - \mathbf{x}_t \rVert + \epsilon},
\end{equation}
where $\epsilon$ is a small constant for numerical stability. A score
of $1$ means the end-effector moves in a perfectly straight line, and
lower scores indicate detours.

\paragraph{Motion Smoothness.}
How steady the motion is, measured by the average magnitude of the
end-effector acceleration (second-order difference), with a negative
sign so that smoother motion receives a higher score:
\begin{equation}
\sigma_{\mathrm{MS}}(\tau) = -\frac{1}{H-2} \sum_{t=2}^{H-1}
\left\lVert \mathbf{x}_{t+1} - 2\mathbf{x}_t + \mathbf{x}_{t-1} \right\rVert .
\end{equation}
Jerky, oscillating motion yields a low score; smooth motion yields a
score close to zero.

\paragraph{Attribute contrast for a trajectory pair.}
Each point in Fig.~\ref{fig:tsne_attribute} corresponds to a trajectory
\emph{pair} $(\tau_A, \tau_B)$, represented by its latent preference
vector $e_{\mathrm{pref}}$. For each attribute $k$, we define the
\emph{behavioral attribute contrast} as
\begin{equation}
\Delta\sigma_k(\tau_A, \tau_B) = \sigma_k(\tau_A) - \sigma_k(\tau_B),
\end{equation}
i.e., which trajectory in the pair scores higher on that attribute and
by how much.

\section{Real-World Experimental Setup}
\label{app:real_world_setup}

This appendix describes the physical setup used in the real-robot
experiments reported in Section~\ref{sec:real_robot}.

\subsection{Hardware and Cameras}
We use a 7-DoF Kinova Jaco2 arm with a two-finger gripper, fixed to the
rear of a tabletop through a clamp-on mount. A black curtain encloses
the scene to keep the background clean. Two cameras observe the
workspace. An Intel RealSense D435 camera is attached to the robot wrist
through a custom bracket, sitting above the gripper and looking at the
area in front of it, which gives a close-up view of the gripper and the
object during execution. A USB camera on a small tripod at the side of
the table provides a fixed view of the whole workspace. Both cameras are
calibrated to the robot with standard procedures. Each metal tray
carries a printed marker on its inner surface for pose estimation, and
object poses are likewise estimated from the calibrated camera streams.
All poses are converted into the same low-dimensional observation format
used in simulation, so the policy sees observations with the same
structure in both domains.

\subsection{Sim-to-Real Configuration}
Policies are trained in Robosuite with its built-in domain randomization
tools~\cite{zhu2020Robosuite}. We randomize system dynamics with
\texttt{DynamicsModder}, covering object mass, friction, joint damping,
and contact parameters, and model realistic sensing with the
\texttt{Observable} interface, which adds sensor noise, sampling-rate
variation, and observation delay. The policy outputs the same actions as
in simulation and sends them to the arm through the manufacturer's
control interface, with motions clipped to a safe region above the
table.

\end{document}